\documentclass[11pt]{article}

\usepackage[preprint]{acl}
\usepackage{enumitem}
\usepackage{times}
\usepackage{latexsym}
\usepackage[table]{xcolor}
\usepackage{booktabs}
\usepackage{amsmath, amssymb}
\usepackage[T1]{fontenc}

\usepackage[utf8]{inputenc}
\usepackage{amsmath}

\usepackage{microtype}

\usepackage{inconsolata}

\usepackage{graphicx}

\title{\textsc{MT-OSC}: Path for LLMs that Get Lost in Multi-Turn Conversation}

\author{
  \textbf{Jyotika Singh},
  \textbf{Fang Tu},
  \textbf{Miguel Ballesteros},
\textbf{Weiyi Sun},
 \\ \textbf{Sandip Ghoshal},
  \textbf{Michelle Yuan},
  \textbf{Yassine Benajiba},
  \textbf{Sujith Ravi},
  \textbf{Dan Roth}
\\
  Oracle AI
\\
  \small{
    \textbf{Correspondence:} \href{mailto:jyotika.s.singh@oracle.com}{jyotika.s.singh@oracle.com}
  }
}

\begin{document}
\maketitle
\begin{abstract}
Large language models (LLMs) suffer significant performance degradation when user instructions and context are distributed over multiple conversational turns, yet multi-turn (MT) interactions dominate chat interfaces. The routine approach of appending full chat history to prompts rapidly exhausts context windows, leading to increased latency, higher computational costs, and diminishing returns as conversations extend.
We introduce \textsc{MT-OSC}, a \textbf{O}ne-off \textbf{S}equential \textbf{C}ondensation framework that efficiently and automatically condenses chat history in the background without disrupting the user experience. MT-OSC employs a Condenser Agent that uses a few-shot inference-based Condenser and a lightweight Decider to selectively retain essential information, reducing token counts by up to 72\% in 10-turn dialogues. Evaluated across 13 state-of-the-art LLMs and diverse multi-turn benchmarks, MT-OSC consistently narrows the multi-turn performance gap—yielding improved or preserved accuracy across datasets while remaining robust to distractors and irrelevant turns.
Our results establish MT-OSC as a scalable solution for multi-turn chats, enabling richer context within constrained input spaces, reducing latency and operational cost, while balancing performance.
\end{abstract}

\section{Introduction}

Large language models (LLMs) have achieved remarkable results on single-turn benchmarks, yet their performance can deteriorate significantly when task instructions or critical context are fragmented across multiple conversational turns—sometimes as early as the second exchange~\citep{laban2025llms, kwan2024mt}. Despite this, such multi-turn (MT) interactions are characteristic of real-world dialogue, where users naturally break up information, clarifications, and intent across several exchanges.

This fragmentation introduces two fundamental challenges: (1) Dispersed information across turns makes it increasingly difficult for LLMs to accurately recall and integrate crucial details, often leading to inconsistencies and degraded performance compared to single-message queries~\citep{laban2025llms}; (2) The prevalent solution of concatenating the entire chat transcript consumes valuable context window and quickly becomes unsustainable, incurring excess latency, cost, and potential information loss as conversations grow.

Although ad-hoc summarization of prior turns has become common, especially for longer chats, these methods risk omitting essential context and are often insufficient for complex multi-turn reasoning. Prior work on chat compression\citep{gliwa2019samsum, bae2022keep, zhang2022history, chen2025compress} and retrieval-augmented dialogues\citep{xu2022beyond, liu2023think} have explored strategies to shrink context size while preserving relevance. Nonetheless, such approaches frequently depend on model fine-tuning, explicit context selection, or complex hierarchical schemes, limiting their scalability and ease of integration.

In contrast, we address a complementary challenge to task decomposition research where prior work improves results by splitting a complex request into multiple focused prompts~\citep{wen-etal-2024-learning, sarangi-etal-2025-decompose, prasad-etal-2024-adapt, WANG2025107200}. Here, we focus on different, realistic scenario where information is \textit{fragmented} over time and must be robustly integrated to avoid compounding errors.

We propose \textsc{MT-OSC}, a \textbf{O}ne-off \textbf{S}equential \textbf{C}ondensation framework that systematically condenses chat history as conversations evolve. MT-OSC operates entirely in the background, requiring no model modifications or fine-tuning and avoiding user-visible latency. Through a combination of a few-shot Condenser and a lightweight, data-driven Decider, MT-OSC dynamically preserves salient context, sharply reducing token overhead and safeguarding against information loss.
Our comprehensive evaluation spanning 13 SOTA LLMs and 10 diverse, multi-turn datasets (spanning underspecified MT samples~\cite{laban2025llms} in sharded BFCL, GSM8K, HumanEval, Spider, ToTTo, and Summary of Haystack, and episodic MT samples from MT-Eval recollection, Expansion, Follow-up, and Refinement benchmarks~\cite{kwan2024mt}) demonstrates that MT-OSC robustly closes the multi-turn performance gap, cutting chat history token significantly and remains resilient even in the presence of synthetically added noise and irrelevant turns. In summary, our primary contributions are:
\begin{itemize}[nosep]
\item A task-agnostic condensing framework (\textbf{MT-OSC}) that achieves up to 72\% \textbf{token reduction} in 10-turn (10 user-assistant exchanges) conversations \textbf{without disrupting user experience through asynchronous background execution}.
\item MT-OSC features a \textbf{Condenser Agent} to carry robust few-shot condensation for effective multi-turn context preservation with a configurable decider to minimize information loss.
\item Extensive experiments across 13 top LLMs and 10 diverse benchmarks confirm MT-OSC’s \textbf{broad applicability and robustness to distractions} and redundant dialogue turns.
\end{itemize}
\section{Related Work}

A central theme in multi-turn research is compressing chat history into compact, reusable state representations instead of naively appending full transcripts~\cite{gliwa2019samsum, feng2021survey, wang2025recursively}. While there have been extensive work on model pre-training~\cite{press2022tstl, su2023rope} and optimization for fine-tuning~\cite{rafailov2024dpo} to improve long context handling, the core challenge remains: preserving task-relevant context as conversations lengthen.
\citet{bae2022keep} tackles this using finetuned T5 summarization model to retain only essential dialogue.
\citet{zhang2022history} leverage hierarchical encodings and attention for iterative refinement of conversation history. 
\citet{chen2025compress} introduce compressive memory by integrating user information with conversation summaries; their approach relies on supervised fine-tuning and preference optimization.
\citet{jiang-etal-2024-longllmlingua} proposed using a small language model to compute token perplexity and selectively remove low-perplexity tokens, also reordering relevant content to beginning or end to address the \emph{lost in the middle}~\cite{lim, levy-etal-2024-task} issue for long-context scenarios; however, this method requires re-compression for each query, limiting the ability to cache context.
In contrast, our approach is distinguished by its lack of reliance on hierarchical encodings, model fine-tuning, heuristic-based compression, or manual context selection. Instead, our method operates unobtrusively in the background, enabling easy integration into any LLM-based chat system without specialized architectures or operational overhead.


Long context prompt compression methods~\cite{pan-etal-2024-llmlingua, jiang-etal-2024-longllmlingua} compress very long inputs via token/segment pruning (often importance/perplexity-based) and filtering/reordering. This differs from short-horizon multi-turn settings, where redundancy is limited and correctness can depend on small but decisive cues (e.g., negations, corrections, key entity mentions); token deletion can therefore alter intent or break coreference, and reordering is less meaningful given the causal chronology of dialogue. As a result, these methods are not directly comparable baselines for MT-OSC, but are better viewed as complementary prompt-compression tools for very long histories.

Another influential line of research, retrieval-augmented conversation~\cite{xu2022beyond, liu2023think, aadhithya2024enhancing, salama-etal-2025-meminsight}, selectively re-injects relevant dialogue snippets using retrieval-based methods to address context growth and improve grounding. While retrieval helps limit prompt size, it may miss nuances spanning multiple turns or temporal dependencies, and often increases system complexity or user latency via real-time retrieval~\cite{chen2025compress}. By contrast, our strategy offers a complementary alternative, maintaining a continuously updated, compact representation that eliminates the need for instantaneous look-up or large-scale context storage.
\begin{figure*}
    \centering
    \includegraphics[width=0.874\linewidth]{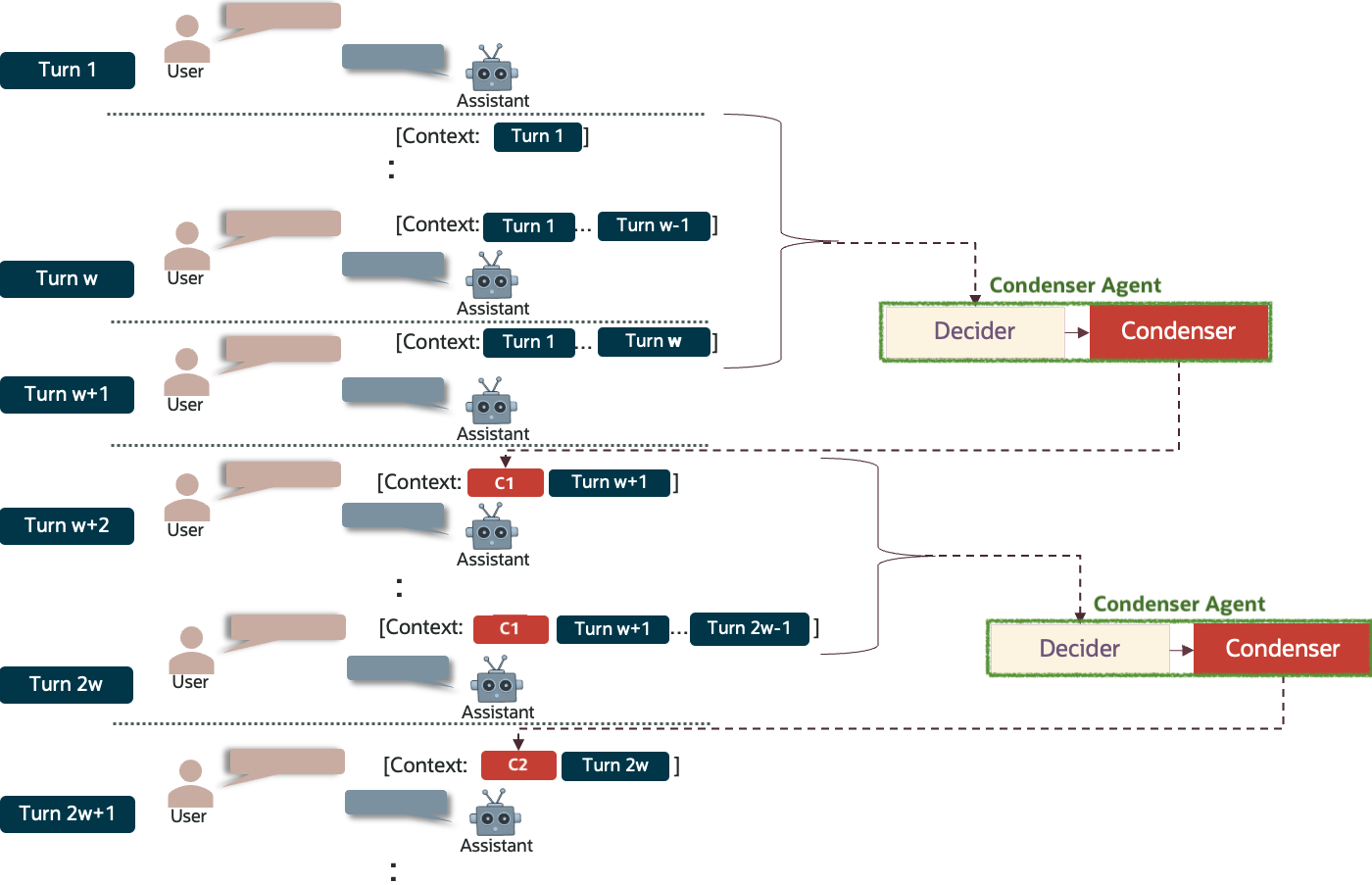}
    \caption{MT-OSC framework with one-off sequential condensation of chat history (w turns/exchanges at a time). This figure illustrates the chat flow when the Condenser Agent initiates condensation of prior history.}
    \label{fig:methodoverview}
\end{figure*}

Recent work has focused on systematic evaluation of multi-turn context handling methods. MT-Eval~\citep{kwan2024mt} benchmarks LLM abilities on dialogue recollection, expansion, refinement, and follow-up. \citet{zhang2025survey} stress evaluation criteria such as conversational quality, multi-turn reasoning, and tool usage. \cite{gorle2025quantifying} presents information loss study as multi-turn conversations progress. \citet{laban2025llms} present a sharding-based evaluation protocol where fully specified single-turn data is artificially split into underspecified multistep interactions to simulate realistic chat dynamics. 
Baseline methods in~\cite{laban2025llms} such as CONCAT and SNOWBALL, although useful for diagnosis, are infeasible for practical deployment: their token requirements scale linearly with chat length, and they often fail for episodic dialogues by repeatedly answering user's older requests. Other proposed methods like RECAP intervene only at the final turn (unknown a priori), limiting their applicability for ongoing dialogues. Furthermore, recommendations such as "retry with more time" or "consolidate before retrying" require human intervention and lack systematic, automated mitigation strategies.

Overall, while prior work has elucidated the scope and severity of multi-turn performance decline, effective and generalizable solutions for automated multi-turn condensation remain relatively underexplored. \textsc{MT-OSC} fills this gap by providing a robust, scalable condensation process that streamlines multi-turn context integration without sacrificing accuracy, latency, or deployability. Appendix~\ref{sec:extrw} shares extended details on related work.

\section{Methods}

MT-OSC Framework (illustrated in Figure~\ref{fig:methodoverview}) comprises a \textit{Condenser agent} comprising a \textit{Condenser} that condenses chat history and a \textit{Decider} which determines the selective application of condensation; and a \textit{one-off sequential} application strategy.

\paragraph{Condenser:}
This module leverages an LLM agent prompted in a few-shot style with explicit reasoning, tuned through curated exemplars to robustly condense chat history. We construct a diverse set of multi-turn chat examples to illustrate a spectrum of condensation policies: cases where key user text should be retained verbatim for future reference, scenarios requiring comprehensive consolidation, and situations where irrelevant turns may be safely excluded. Each hand-crafted example is paired with detailed rationales for information retention. This enables the Condenser to adapt dynamically to distinct dialogue patterns frequently observed in public multi-turn datasets, such as LMSYS-Chat1M~\cite{zheng2024lmsyschatm}. These exemplars are independent of any of the data sources (discussed in Section~\ref{sec:dataset}) we use for our experiments in the paper which will further signify their generality in our results (Section~\ref{sec:results}).

\begin{figure*}
    \centering
    \includegraphics[width=1\linewidth]{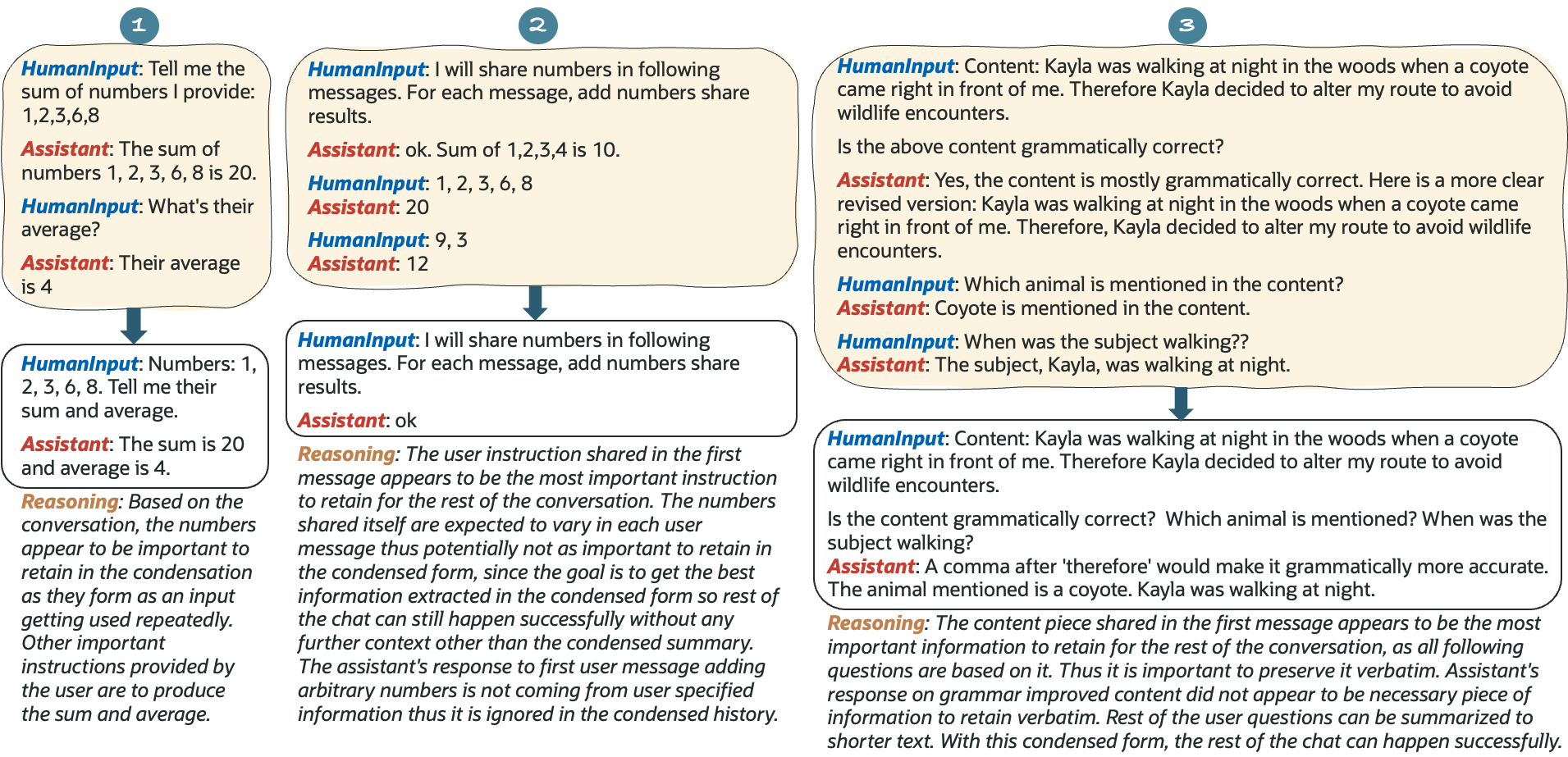}
    \caption{Examples of Multi-Turn Chat Condensing with Reasoning (original chat in yellow; condensed chat below.)}
    \label{fig:fewshots}
\end{figure*}

The few-shot sample set (Figure~\ref{fig:fewshots}) is central to maintaining nuanced context, as our ablation studies demonstrate that generic summarization often misses such subtleties (Section~\ref{sec:results}). Formally:
\[
C_j = \mathrm{Condense}\left( \left\{ (u_i, a_i) \right\}_{i=0}^{w} \right ) = (C_{ju}, C_{ja})
\]

where $C_j$ is the $j$-th condensed user-assistant pair and $w$ denotes the number of turns processed per condensation operation. We use Llama-3.3-70B-Instruct as the underlying Condenser in our experiments; see Appendix~\ref{sec:prompts} for prompt templates.

\paragraph{Decider:}
The Decider is a lightweight and configurable component that adaptively determines whether or not to apply condensation to a given conversation segment. Although condensation substantially reduces token load in many cases, certain dialogues where essential information is repeatedly referenced over several turns with dense context could risk information loss if condensed. To safeguard against this, the Decider employs a rule-based function that quantifies redundancy and novelty in assistant responses relative to prior user queries. Each user message \( U_i \) and assistant response \( A_i \) is represented as a set of normalized content words, following standard preprocessing (stop word removal, case normalization, and lemmatization). For each turn \( i \), we extract unique, relevant words from \( A_i \) and remove those already mentioned in user messages up to turn \( i \) to identify new assistant terms. We further eliminate words that appear in later assistant turns, isolating content unique to each step and tracking overlap. If average overlap between assistant turns exceeds $\gamma$\% and total user content is over $\tau$ tokens across 4 turns, condensation is withheld to avoid potential loss of repeatedly referenced information. This selective approach ensures the condenser maintains fidelity in information-dense dialogues.
Mathematically: 
\begin{align*}
D{w} = 
    \begin{cases}
        1 \quad \begin{aligned}[t]
        \text{if } \mathrm{Overlap}(a_0 -aw)> \gamma~ \text{and}~ \\
        \mathrm{UserTokens}(H_t) > \tau 
        \end{aligned}\\
        0 \quad \text{otherwise}
    \end{cases}
\end{align*}

where Overlap($a_0$, $a_w$) denotes the normalized word overlap between $w$ assistant turns minus user shared content, $\gamma$ is the average overlap threshold (e.g., 0.2), and $\tau$ is the user provided content token threshold in w turns (e.g., 1000).
This approach provides a lightweight, configurable safeguard for information-rich or repetitive conversations. 

In the main paper we share results on set values for Decider hyperparameters ($\gamma$ 0.2, $\tau$ 1000). Appendix~\ref{sec:decsens} contains more details on sensitivity of the Decider parameters. 

\paragraph{One-off Sequential Condensation:}
Relevant information is extracted from prior User and Assistant turns to generate a condensed chat history, which then replaces the raw history. The process condenses the conversation $w$ turns at a time. At turn turn  $w+2$, the condensed history ($C{1}$) is used in place of the original chat history, while the most recent turn ($w+1$) remains in its original form.
The condensation repeats sequentially whenever another $w$ turns have accumulated. ($C{1}$ (computed from the first $w$ turns) plus the next $w-1$ turns after turn $w$). This process repeats as the dialogue progresses, ensuring that only a compact and contextually salient form of history is included in subsequent model inputs.

\noindent\textbf{Combined \textsc{MT-OSC} Operation} is formalized as:
\begin{align*}
H_{w+2} = 
    \begin{cases}
        \left\{ (u_i, a_i) \right\}_{i=0}^{w+1}
         \begin{aligned}[t]
         \quad \textit{(if } D_w \textit{ is False)}
         \end{aligned}\\
        \{(C_{ju}, C_{ja})\} \cup 
        \begin{aligned}[t]
            \{(u_k, a_k)\}_{k = t}^{w+1} \textit{ (else)}
        \end{aligned}
    \end{cases}
\end{align*}

where $t$ is the turn before which the last condensation $C_j$ was performed, $H_w$ is the chat history up to turn $w$, $D_w$ is the output of the Decider for the current window of size $w$, $(u_{i}, a_{i})$ is the user-assistant pair at turn $i$, $C_j$ is the latest condensed representation of the current window, and $w$ is the window size.
See Appendix~\ref{sec:prompts} for more details.

\section{Datasets}
\label{sec:dataset}

We evaluate MT-OSC across 10 datasets, selected to encompass a wide spectrum of multi-turn chat scenarios. These datasets include both sharded static benchmarks and resembling naturally occurring dialogues, representing chats of up to 12 user turns—beyond the multi-turn range typically covered in prior work.

\paragraph{Sharded Datasets.}
We employ six prominent benchmarks~\cite{laban2025llms}—GSM8K, BFCL-V3 Parallel, HumanEval (HEval), Spider, ToTTo, and Summary of Haystack (SoH)—in their sharded form. Each task, originally single-turn, is split into multiple turns to simulate underspecified, real-world multi-turn workflows, where instruction and context are fragmented across user utterances.

\paragraph{MT-EVAL Datasets.}
We further adopt four datasets drawn from MT-EVAL~\cite{kwan2024mt}, each reflecting a different facet of real conversational flow: recollection, refinement, expansion, and follow-up. Noting data scarcity in the recollection and expansion categories (each with only ten instances), we synthesize additional samples for these groups by generating new instances matched to the original style and distribution (see Appendix~\ref{sec:mtevalsyn}).

\paragraph{Dataset Splits.}
We partition all data into development, validation, and test subsets with no instance overlap in downstream evaluation. Seven datasets are held out exclusively for testing; the remainder are split to support model development and hyperparameter tuning. Our primary results are reported on a test set of 752 multi-turn chat samples with 4–12 user turns. Complete breakdowns of corpus sizes and splits are in Appendix~\ref{sec:datasetext}.

\paragraph{Robustness Evaluation Data.}
The datasets are relatively clean without noise, distractions, or unnecessary context. To probe robustness, we inject simulated noise and redundancy between turns to mirror more realistic conversational messiness. These perturbations include: (\emph{i}) \textbf{Repetition Infusion}—randomly duplicating $N$ turns in each chat; (\emph{ii}) \textbf{Filler Injection}—randomly inserting vacuous filler turns (e.g., “Um,” “Hmm”); (\emph{iii}) \textbf{Contextual Diversion}—inserting topically related but task-irrelevant utterances. These methods are inspired by prior attempts to stress-test context robustness~\cite{levy-etal-2024-task, mirzadeh2025gsmsymbolic}, but are newly extended here to multi-turn settings (see Appendix~\ref{sec:promptnoop} for prompt and method details, \ref{sec:perturbedsample} for example of original vs. the three variants).

\paragraph{Evaluation Metrics.}
We use a suite of established metrics tailored to each dataset: accuracy for code and math (HumanEval, BFCL, GSM, Spider, Recollection+); LLM-based quality ratings (10-point scale) for open-ended dialogic tasks (Refinement, Follow-up, Expansion+); BLEU score for ToTTo; and a composite joint-score F1 for SoH. Where prior rule-based metrics proved inadequate (e.g., GSM final answer matching), we augment with LLM-judge reviews for correctness. Full details are in Appendix~\ref{sec:evalappend}.

Unless otherwise stated, our main experiments use Llama-3.3-70B-Instruct. For aggregate performance and token analyses, we restrict to accuracy-based datasets and provide per-task breakdowns as appropriate.

\section{Results and Discussion}
\label{sec:results}
\begin{figure}
    \centering
    \includegraphics[width=1\linewidth]{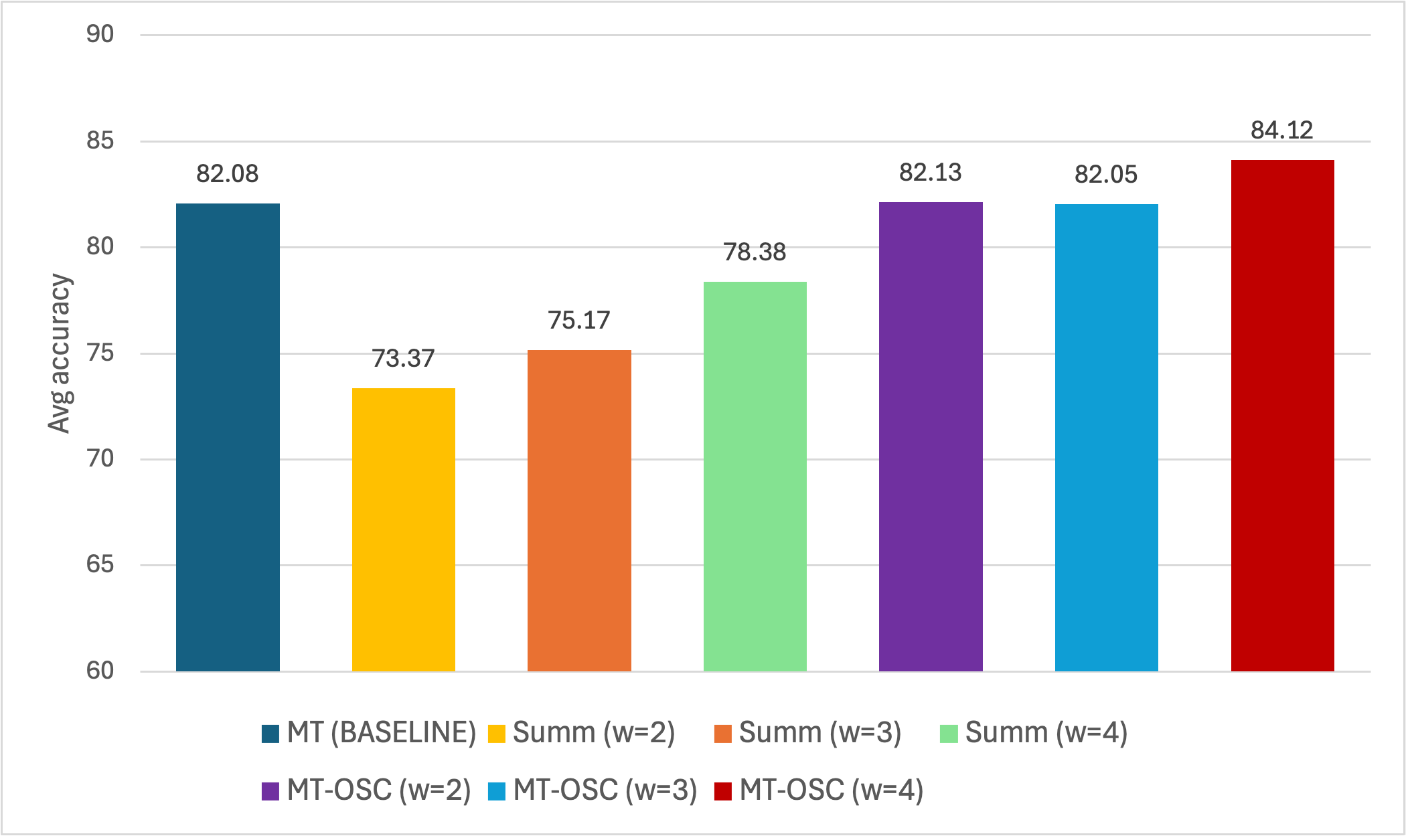}
    \caption{Comparison of performance aggregated across various datasets on simple summarization (Summ) using our one-off sequential application approach and our MT-OSC approach with window sizes 2, 3, 4.}
    \label{fig:resultagg}
\end{figure}
\subsection{Performance vs. Baseline}
We passed each dataset sample through the baseline approach of passing all prior turns into the model's context (MT-baseline) and compare it with MT-OSC at different window sizes, averaged over three+ runs per sample. Figure~\ref{fig:resultagg} shows results across different window sizes, showing MT-OSC with an overall improved performance, with window size 4 showing the most improvement. See Appendix~\ref{sec:resultsext} for details and parameter settings, including information on other baselines (FIFO) that we explored but do not report because their low performance (compared to the industry-standard MT-baseline, which is the default and most widely used approach) makes comparisons uninformative.

Table~\ref{tab:reswdecider} contains the results broken down by dataset and shows that MT-OSC (w=4) achieves improved or largely maintained performance across datasets, while reducing the average chat history tokens.  
Given our exemplars are from an independent source than the datasets we present results on; this supports their generality in our method.
Statistical analysis via Wilcoxon signed-rank test shows a \textbf{\textit{statistically significant reduction in token count}} ($statistic=28.0$, $p=0.009$), while performance remains statistically indistinguishable ($statistic=7.0$, $p=0.118$) from the MT-baseline—demonstrating MT-OSC’s efficiency without compromising on performance. 

\begin{table}[h!]
\centering
\begin{tabular}{p{1.43cm}|p{1.035cm}|p{0.965cm}|p{1.035cm}|p{0.9cm}}

 & \multicolumn{2}{p{2.0cm}|}{Performance} & \multicolumn{2}{p{1.935cm}}{Chat history tokens} \\
  & MT-baseline & MT-OSC* & MT-baseline & MT-OSC*\\
\hline
\hline

BFCL$^{shrd}$ & 81.13\%& \textbf{86.79}\% & 165 & \textbf{124} \\
GSM$^{shrd}$ & 83.45\% & \textbf{84.80}\% & 1260 & \textbf{354}\\
HEval$^{shrd}$ & 74.67\% & \textbf{77.33}\% & 539 & \textbf{272}\\
Spider$^{shrd}$ & 76.95\% & \textbf{79.44}\% & 105 & 104 \\
ToTTo$^{shrd}$ & 0.19 & 0.19 & 2585 & 2585\\
SoH$^{shrd}$ & 0.13 & 0.13 & 7495 & 7495 \\
refine$^{mte}$ & 5.39 & 5.35 & 1663 & \textbf{1342}\\
follow$^{mte}$ & 8.47 & 8.47 & 1379 & 1379\\
recoll$^{mte+}$ & 94.22\% & 92.22\%  & 1258 & \textbf{562}\\
expan$^{mte+}$ & 8.61 & 7.9 & 873 & \textbf{648} \\
\hline
\end{tabular}
\caption{Average performance and total chat history tokens (per turn) comparing MT-baseline and MT-OSC (w=4) across datasets. Metrics: accuracy (BFCL, GSM, HEval, Spider, recoll), 10-point rating (refine, follow, expan), BLEU (ToTTo), and joint score (SoH).}
\label{tab:reswdecider}
\end{table}

\paragraph{Context Loss Mitigation:} The common reasons for drop in performance in multi-turn conversations is the risk of either failing to recall pertinent context or propagating incorrect assumptions—especially before users provide sufficient detail or clarification \cite{laban2025llms}. This issue is particularly prominent in sharded datasets such as GSM, BFCL, HumanEval, and Spider, where initial conversational turns frequently lack adequate context, causing large language models to make premature or unsupported inferences. As subsequent turns unfold and more information becomes available, models sometimes struggle to reconcile earlier assumptions, potentially leading to confusion or compounded errors.
Notably, in these challenging scenarios, MT-OSC offers a clear advantage by condensing chat history to include only the most salient and contextually important information leading to improved accuracy and more consistent reasoning, as demonstrated by our quantitative results. For illustrative cases where this approach led to correction of earlier misinterpretations, we provide detailed examples in Appendix~\ref{sec:examples} (Figure~\ref{fig:example}).


\subsection{Token Reduction} 
\begin{figure}
    \centering
    \includegraphics[width=1\linewidth]{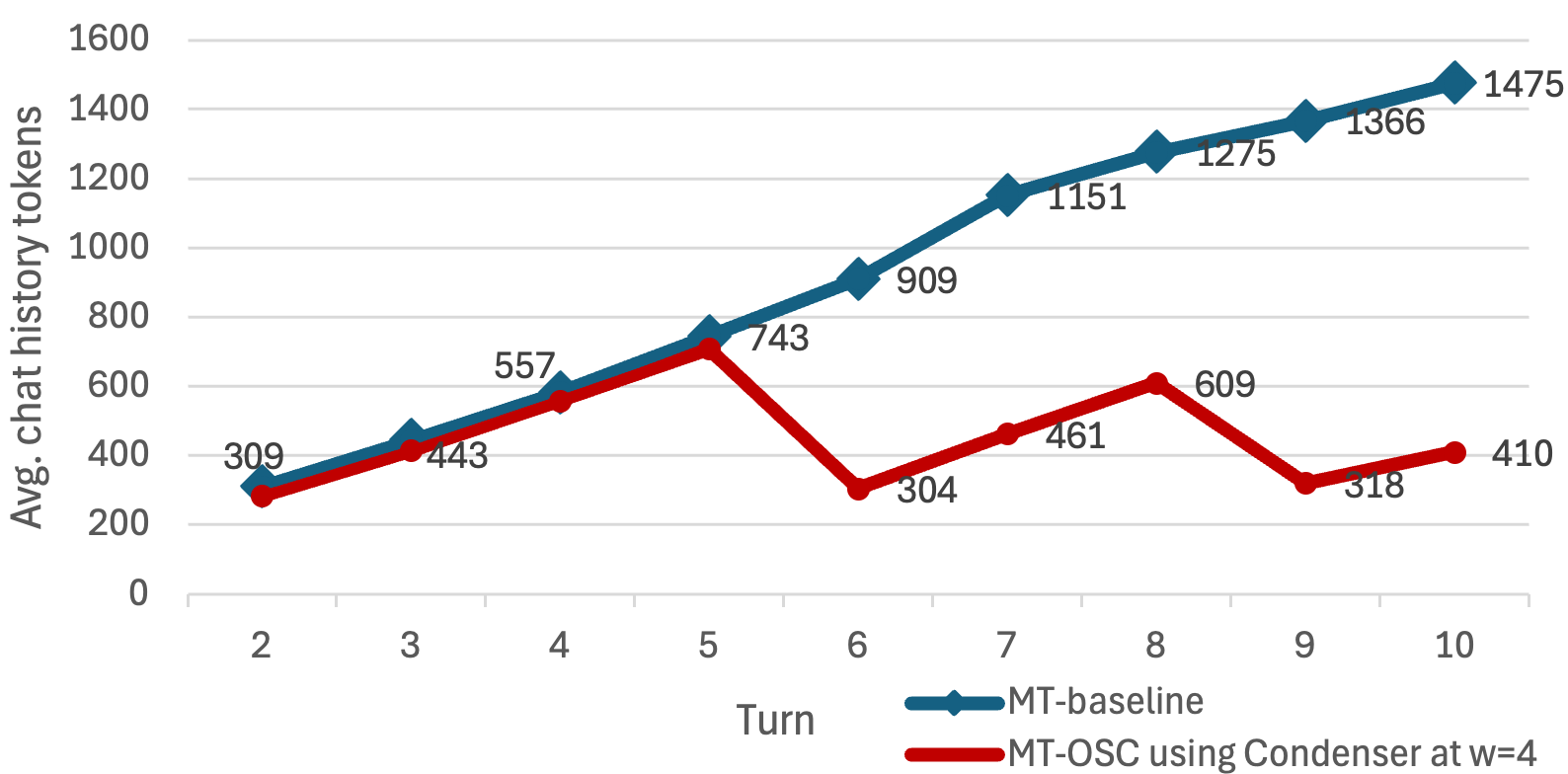}
    \caption{Chat history tokens growth in MT-baseline compared to using Condenser ($w=4$) showing reduction in token consumption in the user-facing chat.
}
    \label{fig:tokenred}
\end{figure}

\begin{figure*}
    \centering
    \includegraphics[width=1.0\linewidth]{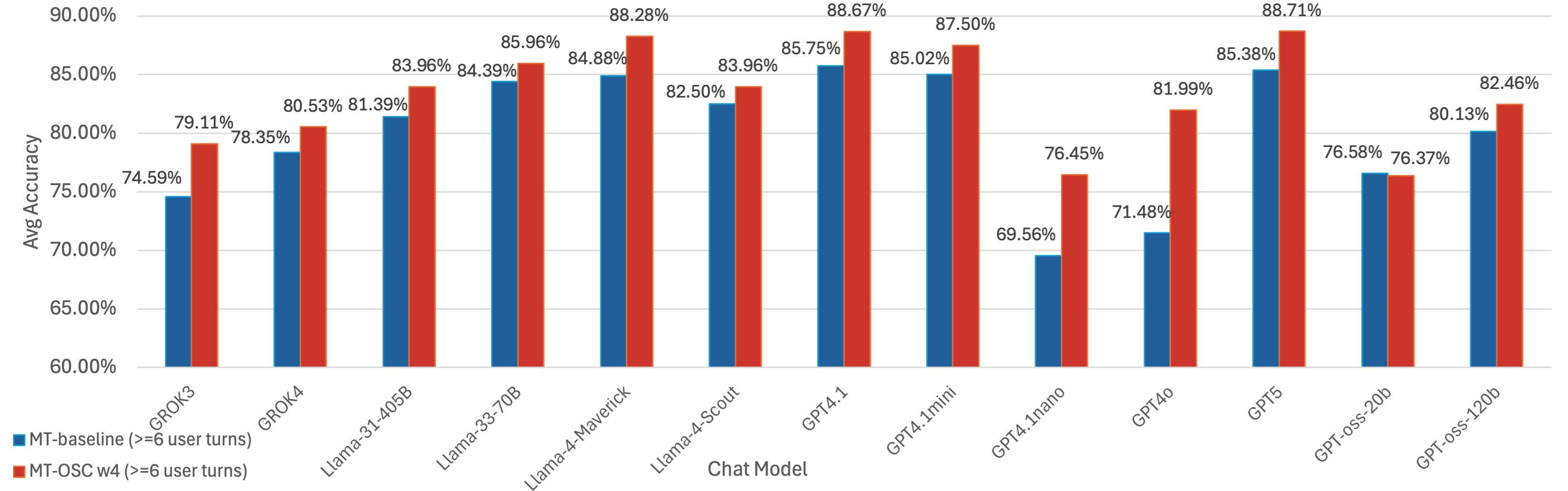}
    \caption{Model performance scores for various Chat Models for conversations with 6 or more turns across baseline multi-turn and MT-OSC with window size $w=4$. See Table~\ref{tab:llmsdatasetsres} for broken down details in tabular form.}
    \label{fig:llms}
\end{figure*}

Figure~\ref{fig:tokenred} shows that \textbf{\textit{MT-OSC significantly reduces chat history tokens up to 72\% for 10-turn chats}} compared to passage of full chat history to the model's context in MT-baseline, \textbf{\textit{bypassing linear growth in token consumption as the chat grows.}}

Table~\ref{tab:reswdecider} presents the results broken down by dataset and shows that MT-OSC (\(w=4\)) reduces the average number of chat history tokens. The only datasets where token counts do not decrease are those in which most samples have fewer than \(w+2\) total turns (e.g., MT-EVAL-follow-up and Spider (with only 6 samples)), or where the Decider did not pass the samples through the Condenser component (e.g., ToTTo and SoH). Other datasets that utilized the Condenser show reductions in chat history tokens ranging from 19\% to 72\%, with an average reduction of 41\%. These results include both samples with fewer than six turns and those with six or more turns. The true impact of the Condenser is observed in conversations with six or more turns, where chat history tokens are reduced by an average of 44.56\% across those datasets, as shown in Figure~\ref{fig:tokencomp} in Appendix~\ref{sec:decider}.

\paragraph{Chat Efficiency for the End-User:} MT-OSC reduces the chat history included in the main chat prompt, thereby decreasing the number of tokens and consequently lowering latency for the end user. Prompt length is a primary contributor to Time To First Token (TTFT), which measures how quickly users begin to see the model’s output \cite{databricks}. Empirical studies have shown a linear relationship between the number of prompt tokens and TTFT \cite{glean}. \textbf{\textit{For chats with >=6 turns, if chat history comprises 4,000 tokens, MT-OSC saves an average of 1,782 tokens per exchange}} (reducing from 4,000 to 2,218 tokens), which translates to \textit{\textbf{approximately 1.1-1.2 seconds in latency reduction}} based on findings in \cite{glean}. This reduction provides a tangible improvement in end-user experience.

In real-world applications, the number of turns persisted in a chat may often be restricted to conserve computational resources, with a limit of around 10 turns for many exploratory dialogues~\cite{localiq,blockgit}. MT-OSC maintains low chat history token consumption, limiting history turns to <=5 when w=4, thereby enabling efficient operation within these practical constraints.

\begin{figure}[]
    \centering
    \includegraphics[width=1\linewidth]{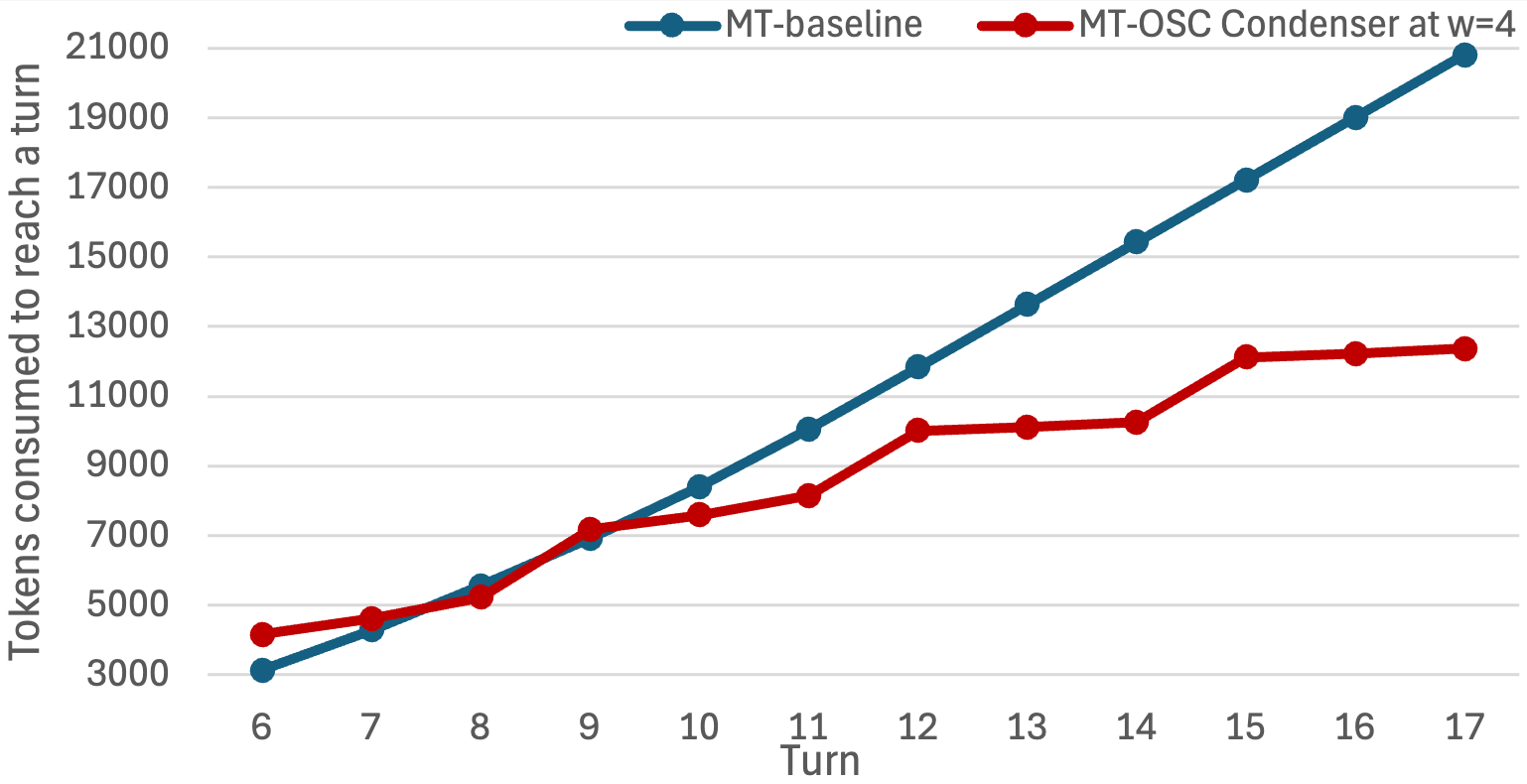}
    \caption{Total tokens consumed by chat history in the main chat + background condenser, showing overall cost benefit and overhead balancing as the chat grows.}
    \label{fig:tottokens}
\end{figure}

\paragraph{Overall Solution Efficiency (User-Facing Chat + Background Process):} Tokens are still consumed during the condensing operation, which is processed in the background and does not affect the user experience. This background processing nevertheless contributes to the overall token usage and associated costs. To quantify the balance between token savings and additional background consumption, Figure~\ref{fig:tottokens} presents the total number of tokens used—encompassing both real-time chat and background condenser operations—for the baseline multi-turn scenario and for MT-OSC Condenser. In comparison, \textbf{\textit{as conversations progress, the overall token consumption is a lot lower with condensation}}, highlighting the efficiency of our approach.

\paragraph{Asynchronous Execution Advantage}
MT-OSC runs as a background process whose output is not required for the immediate next turn in the chat. Since its prompt length is typically comparable to or smaller than the main chat—often shorter due to additional system prompts and user typing delays—the condensation step usually completes while the next turn wraps. Its result is therefore available by the next-to-next turn.
This enables a one-off, asynchronous condensation step that \textbf{\textit{adds \emph{no latency} to the user experience}}: the conversation proceeds normally while MT-OSC executes in the background, and its output is seamlessly used in one-off subsequent turns.

\subsection{Performance Across LLMs}

\paragraph{Generalization Across Chat LLMs}
We evaluated MT-OSC with a window size w=4 across 13 leading LLMs compared to MT-baseline, as summarized in Figure~\ref{fig:llms}. 
\textbf{\textit{Consistently across models, including the latest releases, MT-OSC outperformed the baseline in task accuracy}}. See Appendix~\ref{sec:resultsext} for detailed results by dataset.

Notably, GPT-5 achieved the highest performance under the MT-baseline among the tested models followed by GPT4o, yet MT-OSC further boosted its performance by 3.3\%. Similarly, we notice improvement with MT-OSC across Llama and Grok models. These results demonstrate that MT-OSC delivers consistent gains in performance across a diverse range of top models across different LLM architectures.

\paragraph{Generalization Across Condenser LLMs}
In this paper main experiments, we report results using Llama-3.3-70B as the fixed Condenser LLM. To assess whether MT-OSC generalizes across Condenser models, we additionally reran a subset of datasets (Sharded BFCL and HumanEval) with alternative Condenser models. As shown in Table~\ref{tab:condmodelres} (Appendix~\ref{sec:condmodelres}), replacing the Condenser model with Llama-4-Maverick, GPT-4.1, or Gemini-2.5-Flash—while keeping the overall setup, prompts, and hyperparameters unchanged—\textbf{yields consistent performance retention or improvement} on the evaluated subset.

\subsection{Ablation Study}
To better understand the contribution of individual components in MT-OSC, we conduct two targeted ablation studies: 1) \textbf{Decider:} We evaluate the impact of the Decider by comparing full MT-OSC to a variant of the Condenser Agent with the Decider removed, measuring performance differences. 2) \textbf{Condenser vs. Simple Summarization:} With Decider disabled and $w=$ (2, 3, 4), we replace our Condenser module with a simple summarization prompt (see Appendix~\ref{sec:prompts} for details), which provides pointed summarization instructions but omits the curated examples used in our Condenser.


\begin{table}[h!]
\centering
\begin{tabular}{p{1.405cm}|p{0.9cm}|p{0.92cm}||p{1.1cm}|p{1cm}}
 & \multicolumn{2}{p{2cm}||}{Performance} & \multicolumn{2}{p{2cm}}{Chat Hist Tokens} \\
Dataset  & w/ D & w/o D & w/ D & w/o D \\
\hline
\hline
ToTTo$^{shrd}$ & \textbf{0.18} & 0.09 & 2585 & \textbf{1888}\\
SoH$^{shrd}$ & \textbf{0.13} & 0.08 & 7495 & \textbf{4086}\\
Refin$^{mtev}$ & \textbf{5.35} & 5.18 & 1342 & \textbf{593}\\
\hline
\end{tabular}
\caption{Performance and chat history tokens with and without Decider (D) across the datasets where Decider induces a performance or token count difference.}
\label{tab:deciderabl}
\end{table}
\paragraph{Decider Ablation:}Table~\ref{tab:deciderabl} summarizes the results on datasets that led to a performance difference without the Decider component. We find that the Decider component consistently maintains high chat success rates, especially on more challenging datasets where condensation may likely lose relevant information. \textbf{\textit{Ablation of the Decider generally leads to reduced performance on three datasets, highlighting its role in safeguarding against information loss during condensation}}. Since w=4 shows an impact on chats with six or more turns, we compute the reduction in chat history tokens per dataset for chats with at least six turns. \textbf{\textit{With the Decider, chat history tokens are reduced by an average of 44.56\%, and without the Decider, the reduction is 52.07\%}} (Figure~\ref{fig:tokencomp} in Appendix~\ref{sec:decider}). 
Decider is a lightweight module and tunable based on the nature of the data and priority between performance and token length. An example from the Sharded ToTTo dataset where Decider skips condensing and prevents information loss is shared in Appendix~\ref{sec:decider}, and details on sensitivity of the Decider parameters are shared in Appendix~\ref{sec:decsens}. 

\paragraph{Condenser vs. Simple Summarization:} Figure~\ref{fig:resultagg} shows that replacing the Condenser with a \textbf{\textit{simple summarizer decreases success rates across datasets and window sizes}} demonstrating the effectiveness of our curated few-shot approach in maintaining better shortened chat histories and the limitations of the summary prompt baseline. 

In our experiments, we observe that the summary prompt often omits important user instructions during summarization. In contrast, our condenser-based method provides critical examples, enabling the model to preserve the full text when necessary and to condense or omit less relevant information as appropriate. Notably, our approach achieves these improvements without reliance on a reasoning-specialized LLMs given the demonstrated performance using Llama-3.3-70B-Ins as the Condenser model. An example showing Condenser and simple summarizer outputs is shared in Appendix~\ref{sec:examples}.

\noindent{}These studies show that the Condenser is essential to MT-OSC’s ability to preserve and often enhance performance in multi-turn dialogue settings.  The Decider, as a lightweight and tunable module, offers additional protection against performance degradation, especially on complex or information-dense datasets, by safeguarding against potential information loss during condensation.


\subsection{Robustness Testing}
\begin{figure}
    \centering
    \includegraphics[width=1\linewidth]{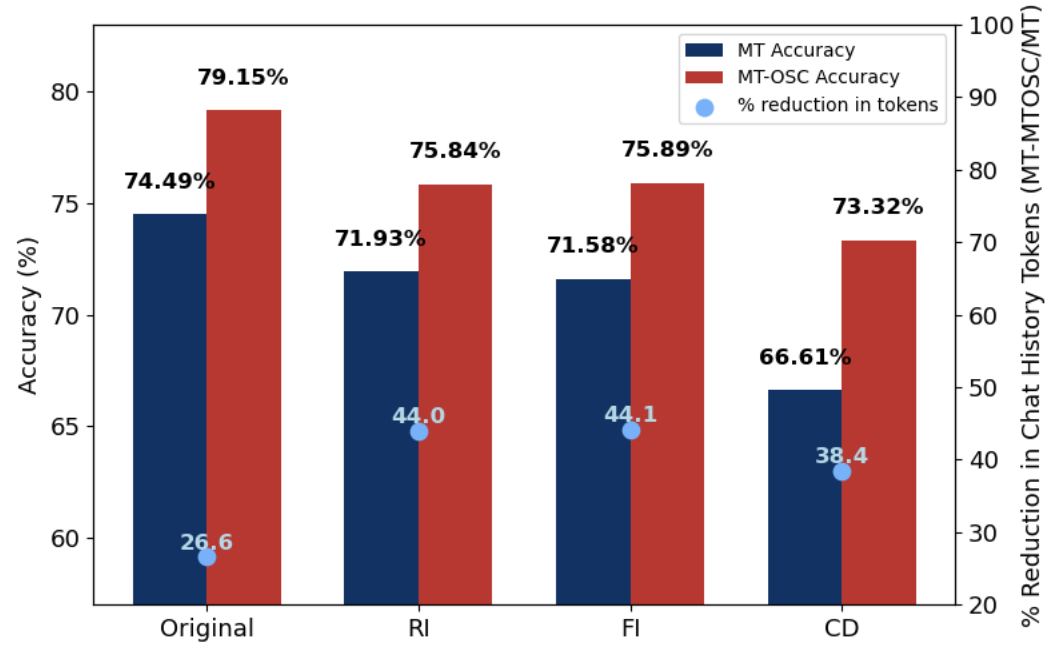}
    \caption{Comparison of performance between MT and MT-OSC, and reduction in chat history tokens, on the three different variants created for robustness testing (1. Repetition Infusion (RI); 2. Filler Injection (FI); 3. Contextual Diversion (CD)).}
    \label{fig:rob}
\end{figure}
We evaluated the robustness of our approach using three simulated variants of the original datasets (as described in Section~\ref{sec:dataset}) with 1-2 added turns, applying both full multi-turn chat execution and MT-OSC. The comparative results are presented in Figure~\ref{fig:rob} aggregated across Grok-3, Llama-3.3-70B, Llama-4-Maverick, Llama-4-Scout, GPT4.1, and GPT4o and sharded BFCL, GSM, HEval, and Spider datasets. See Appendix~\ref{sec:robustresults} for detailed results by dataset.



Our results show that the \textit{\textbf{inclusion of extra turns degrades multi-turn performance relative to the clean version of the data}}, with a performance drop of about 7.88\% for the contextual diversion variant and 2.5-3\% for the other two variants (see Figure~\ref{fig:rob}). This drop aligns with prior work, which reports that increased context length can impair task performance~\cite{levy-etal-2024-task}. Our results extend this observation to multi-turn scenarios.


The results show that \textbf{\textit{MT-OSC works just as well on these variants showing similar (about 4\%) or even higher (up to 7\% in Contextual Diversion variants) absolute gains in performance than on original/cleaner data while reducing chat history tokens}}. Importantly, MT-OSC effectively mitigates the impact of unnecessary information by condensing the chat history to retain only the most relevant content, maintaining stable performance benefits even on noisier data.

\begin{figure}
    \centering
    \includegraphics[width=1\linewidth]{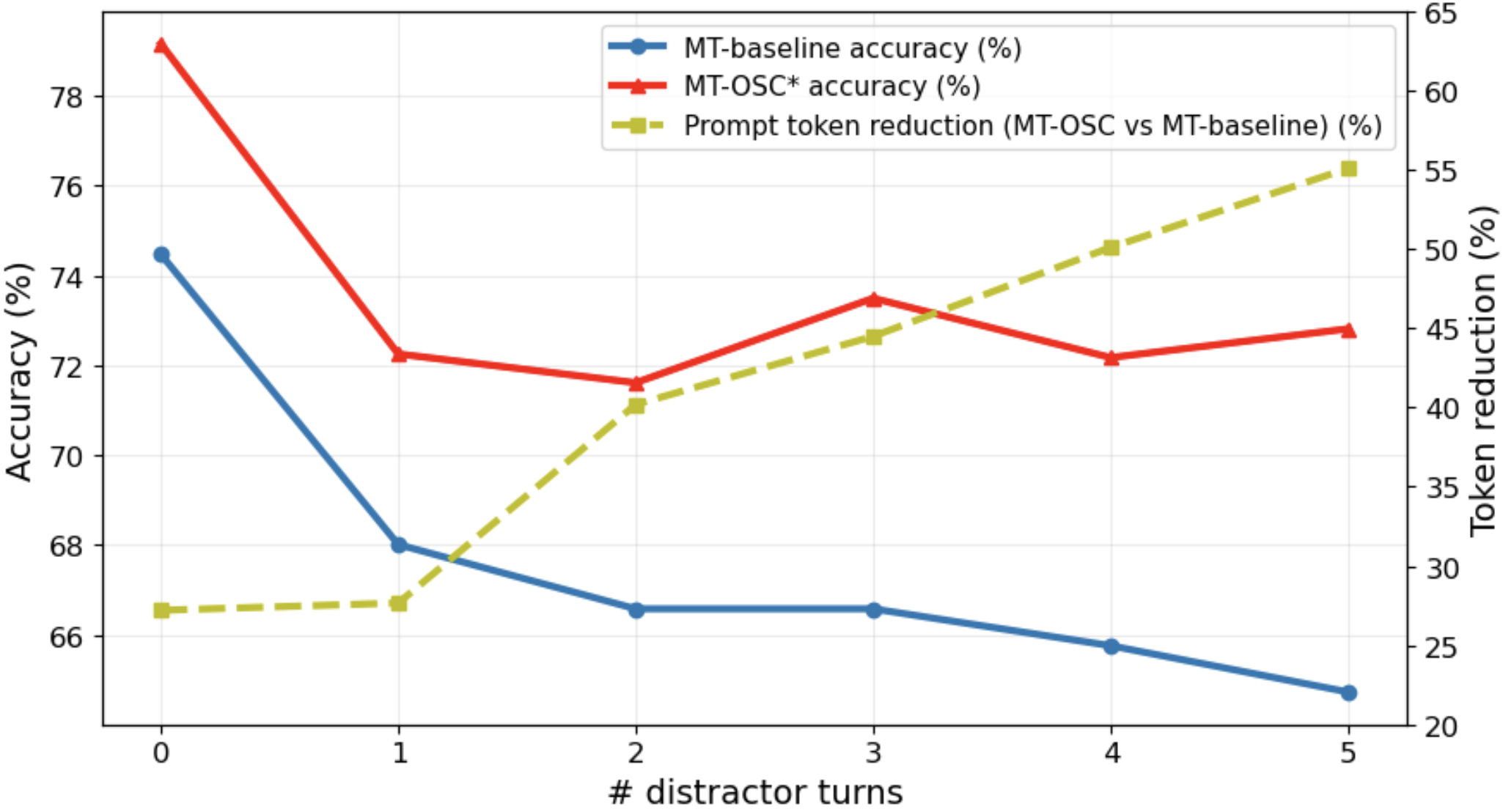}
    \caption{Performance of MT-baseline and MT-OSC under Contextual Diversion (CD) variants varied by number of distraction turns added.}
    \label{fig:rob1}
\end{figure}

To test \textbf{robustness against varying amounts of noise}, we also conducted an experiment to vary the number of extra turns added (1-5) for the contextual divergent variants. Figure~\ref{fig:rob1} shows that increasing the number of distractor turns (and thus the conversation length and user input tokens) degrades baseline performance and MT-OSC shows consistent improvements throughout. \textbf{\textit{Token reduction increases as the amount of distracting content grows}}, showing that \textbf{\textit{MT-OSC successfully handles noise}} and retains task-relevant information.

\section{Conclusion}

Multi-turn conversations present persistent challenges for LLMs, resulting in both degraded task performance and unsustainable increases in prompt length as dialogues scale. To address these limitations, we introduce MT-OSC, background condensation framework that sequentially condenses chat history as conversations grow. MT-OSC achieves significant reductions in chat history token counts (up to 72\% fewer tokens for 10-turn dialogues) while either preserving or enhancing accuracy across a diverse set of benchmarks and state-of-the-art LLM architectures.

Our evaluation demonstrates that MT-OSC not only narrows the multi-turn performance gap but also provides robust gains in settings with conversational noise or distractors, enabling more efficient and reliable interactions. Importantly, MT-OSC operates entirely in the background: it does not require model modification, retraining, or user intervention, and it delivers tangible improvements in latency and efficiency for real-world deployments.

Overall, MT-OSC provides a practical, scalable solution to the multi-turn context bottleneck, enabling broader applications of conversational AI in chat systems.

\section*{Limitations}

This study utilizes publicly available datasets that are designed for comparing single-turn and multi-turn conversational interactions. These datasets typically contain fewer than 10–12 user turns per conversation and primarily feature clean, single-topic dialogues with minimal agentic behavior—that is, they involve straightforward LLM calls without exploring the tool-use capabilities that confer agentic properties to LLMs. While we applied data augmentation to assess robustness, the underlying data represents limitation in linguistic complexity and topic diversity that characterize more diverse real-world conversational settings.


For \textbf{future work}, model training can explicitly incorporate more complex multi-turn interactions, enabling forthcoming models to better handle the challenges presented by extended conversations.
Developing or incorporating datasets with longer, more diverse, and multi-topic conversations would provide a richer foundation for analysis and modeling. Further, leveraging data that reflects greater conversational complexity, including open-domain and agent-driven exchanges, could enhance the generalizability and applicability of future findings. Exploring methods such as topical decomposition may also provide valuable support for tackling the challenges posed by complex, multi-faceted dialogue interactions.

\bibliography{custom}

\begin{thebibliography}{54}
\providecommand{\natexlab}[1]{#1}

\bibitem[{Aadhithya et~al.(2024)Aadhithya, Kumar, and Soman}]{aadhithya2024enhancing}
Aadharsh Aadhithya, Sachin Kumar, and KP~Soman. 2024.
\newblock \href {https://arxiv.org/abs/2406.06124} {Enhancing long-term memory using hierarchical aggregate tree for retrieval augmented generation}.
\newblock \emph{arXiv e-prints}, pages arXiv--2406.

\bibitem[{Agarwal et~al.(2025)Agarwal, Meghwani, Patel, Sheng, Ravi, and Roth}]{agarwal2025aligning}
Amit Agarwal, Hansa Meghwani, Hitesh~Laxmichand Patel, Tao Sheng, Sujith Ravi, and Dan Roth. 2025.
\newblock \href {https://doi.org/10.18653/v1/2025.emnlp-industry.9} {Aligning {LLM}s for multilingual consistency in enterprise applications}.
\newblock In \emph{Proceedings of the 2025 Conference on Empirical Methods in Natural Language Processing: Industry Track}, pages 117--137, Suzhou (China). Association for Computational Linguistics.

\bibitem[{Agarwal et~al.(2023)Agarwal, Qureshi, Sardana, Li, Quevedo, and Khudia}]{databricks}
Megha Agarwal, Asfandyar Qureshi, Nikhil Sardana, Linden Li, Julian Quevedo, and Daya Khudia. 2023.
\newblock \href {https://www.databricks.com/blog/llm-inference-performance-engineering-best-practices} {Llm inference performance engineering: Best practices}.

\bibitem[{Bae et~al.(2022)Bae, Kwak, Kang, Lee, Kim, Jeong, Kim, Lee, Park, and Sung}]{bae2022keep}
Sanghwan Bae, Donghyun Kwak, Soyoung Kang, Min~Young Lee, Sungdong Kim, Yuin Jeong, Hyeri Kim, Sang-Woo Lee, Woomyoung Park, and Nako Sung. 2022.
\newblock \href {https://doi.org/10.18653/v1/2022.findings-emnlp.276} {Keep me updated! memory management in long-term conversations}.
\newblock In \emph{Findings of the Association for Computational Linguistics: EMNLP 2022}, pages 3769--3787, Abu Dhabi, United Arab Emirates. Association for Computational Linguistics.

\bibitem[{Bai et~al.(2024)Bai, Liu, Bu, He, Liu, Zhou, Lin, Su, Ge, Zheng, and Ouyang}]{bai-etal-2024-mt}
Ge~Bai, Jie Liu, Xingyuan Bu, Yancheng He, Jiaheng Liu, Zhanhui Zhou, Zhuoran Lin, Wenbo Su, Tiezheng Ge, Bo~Zheng, and Wanli Ouyang. 2024.
\newblock \href {https://doi.org/10.18653/v1/2024.acl-long.401} {{MT}-bench-101: A fine-grained benchmark for evaluating large language models in multi-turn dialogues}.
\newblock In \emph{Proceedings of the 62nd Annual Meeting of the Association for Computational Linguistics (Volume 1: Long Papers)}, pages 7421--7454, Bangkok, Thailand. Association for Computational Linguistics.

\bibitem[{Chen et~al.(2021)Chen, Tworek, Jun, Yuan, Pond{\'e}, Kaplan, Edwards, Burda, Joseph, Brockman, Ray, Puri, Krueger, Petrov, Khlaaf, Sastry, Mishkin, Chan, Gray, Ryder, Pavlov, Power, Kaiser, Bavarian, Winter, Tillet, Such, Cummings, Plappert, Chantzis, Barnes, Herbert-Voss, Guss, Nichol, Babuschkin, Balaji, Jain, Carr, Leike, Achiam, Misra, Morikawa, Radford, Knight, Brundage, Murati, Mayer, Welinder, McGrew, Amodei, McCandlish, Sutskever, and Zaremba}]{Chen2021EvaluatingLL}
Mark Chen, Jerry Tworek, Heewoo Jun, Qiming Yuan, Henrique Pond{\'e}, Jared Kaplan, Harrison Edwards, Yura Burda, Nicholas Joseph, Greg Brockman, Alex Ray, Raul Puri, Gretchen Krueger, Michael Petrov, Heidy Khlaaf, Girish Sastry, Pamela Mishkin, Brooke Chan, Scott Gray, and 34 others. 2021.
\newblock \href {https://api.semanticscholar.org/CorpusID:235755472} {Evaluating large language models trained on code}.
\newblock \emph{ArXiv}, abs/2107.03374.

\bibitem[{Chen et~al.(2025)Chen, Li, Chang, Huang, Wang, and Li}]{chen2025compress}
Nuo Chen, Hongguang Li, Jianhui Chang, Juhua Huang, Baoyuan Wang, and Jia Li. 2025.
\newblock \href {https://aclanthology.org/2025.coling-main.51/} {Compress to impress: Unleashing the potential of compressive memory in real-world long-term conversations}.
\newblock In \emph{Proceedings of the 31st International Conference on Computational Linguistics}, pages 755--773, Abu Dhabi, UAE. Association for Computational Linguistics.

\bibitem[{Cobbe et~al.(2021)Cobbe, Kosaraju, Bavarian, Chen, Jun, Kaiser, Plappert, Tworek, Hilton, Nakano, Hesse, and Schulman}]{Cobbe2021TrainingVT}
Karl Cobbe, Vineet Kosaraju, Mo~Bavarian, Mark Chen, Heewoo Jun, Lukasz Kaiser, Matthias Plappert, Jerry Tworek, Jacob Hilton, Reiichiro Nakano, Christopher Hesse, and John Schulman. 2021.
\newblock \href {https://api.semanticscholar.org/CorpusID:239998651} {Training verifiers to solve math word problems}.
\newblock \emph{ArXiv}, abs/2110.14168.

\bibitem[{Duan et~al.(2024)Duan, Wei, Wang, Liu, Fang, Zhang, Lin, and Chen}]{duan-etal-2024-botchat}
Haodong Duan, Jueqi Wei, Chonghua Wang, Hongwei Liu, Yixiao Fang, Songyang Zhang, Dahua Lin, and Kai Chen. 2024.
\newblock \href {https://doi.org/10.18653/v1/2024.findings-naacl.201} {{B}ot{C}hat: Evaluating {LLM}s' capabilities of having multi-turn dialogues}.
\newblock In \emph{Findings of the Association for Computational Linguistics: NAACL 2024}, pages 3184--3200, Mexico City, Mexico. Association for Computational Linguistics.

\bibitem[{Feng et~al.(2022)Feng, Feng, and Qin}]{feng2021survey}
Xiachong Feng, Xiaocheng Feng, and Bing Qin. 2022.
\newblock \href {https://arxiv.org/abs/2107.03175} {A survey on dialogue summarization: Recent advances and new frontiers}.
\newblock \emph{arXiv preprint arXiv:2107.03175}.

\bibitem[{Gliwa et~al.(2019)Gliwa, Mochol, Biesek, and Wawer}]{gliwa2019samsum}
Bogdan Gliwa, Iwona Mochol, Maciej Biesek, and Aleksander Wawer. 2019.
\newblock \href {https://doi.org/10.18653/v1/D19-5409} {{SAMS}um corpus: A human-annotated dialogue dataset for abstractive summarization}.
\newblock In \emph{Proceedings of the 2nd Workshop on New Frontiers in Summarization}, pages 70--79, Hong Kong, China. Association for Computational Linguistics.

\bibitem[{Gorle et~al.(2025)Gorle, Yadav, and Weissman}]{gorle2025quantifying}
Abhiram~Rao Gorle, Amit Kumar~Singh Yadav, and Tsachy Weissman. 2025.
\newblock \href {https://openreview.net/forum?id=5gpABTkcUJ} {Quantifying information gain and redundancy in multi-turn {LLM} conversations}.
\newblock In \emph{First Workshop on Multi-Turn Interactions in Large Language Models}.

\bibitem[{Jiang et~al.(2024)Jiang, Wu, Luo, Li, Lin, Yang, and Qiu}]{jiang-etal-2024-longllmlingua}
Huiqiang Jiang, Qianhui Wu, Xufang Luo, Dongsheng Li, Chin-Yew Lin, Yuqing Yang, and Lili Qiu. 2024.
\newblock \href {https://doi.org/10.18653/v1/2024.acl-long.91} {{L}ong{LLML}ingua: Accelerating and enhancing {LLM}s in long context scenarios via prompt compression}.
\newblock In \emph{Proceedings of the 62nd Annual Meeting of the Association for Computational Linguistics (Volume 1: Long Papers)}, pages 1658--1677, Bangkok, Thailand. Association for Computational Linguistics.

\bibitem[{Kwan et~al.(2024)Kwan, Zeng, Jiang, Wang, Li, Shang, Jiang, Liu, and Wong}]{kwan2024mt}
Wai-Chung Kwan, Xingshan Zeng, Yuxin Jiang, Yufei Wang, Liangyou Li, Lifeng Shang, Xin Jiang, Qun Liu, and Kam-Fai Wong. 2024.
\newblock \href {https://doi.org/10.18653/v1/2024.emnlp-main.1124} {{MT}-eval: A multi-turn capabilities evaluation benchmark for large language models}.
\newblock In \emph{Proceedings of the 2024 Conference on Empirical Methods in Natural Language Processing}, pages 20153--20177, Miami, Florida, USA. Association for Computational Linguistics.

\bibitem[{Laban et~al.(2024)Laban, Fabbri, Xiong, and Wu}]{laban-etal-2024-summary}
Philippe Laban, Alexander Fabbri, Caiming Xiong, and Chien-Sheng Wu. 2024.
\newblock \href {https://doi.org/10.18653/v1/2024.emnlp-main.552} {Summary of a haystack: A challenge to long-context {LLM}s and {RAG} systems}.
\newblock In \emph{Proceedings of the 2024 Conference on Empirical Methods in Natural Language Processing}, pages 9885--9903, Miami, Florida, USA. Association for Computational Linguistics.

\bibitem[{Laban et~al.(2025)Laban, Hayashi, Zhou, and Neville}]{laban2025llms}
Philippe Laban, Hiroaki Hayashi, Yingbo Zhou, and Jennifer Neville. 2025.
\newblock \href {https://arxiv.org/abs/2505.06120} {Llms get lost in multi-turn conversation}.
\newblock \emph{Preprint}, arXiv:2505.06120.

\bibitem[{Levy et~al.(2024)Levy, Jacoby, and Goldberg}]{levy-etal-2024-task}
Mosh Levy, Alon Jacoby, and Yoav Goldberg. 2024.
\newblock \href {https://doi.org/10.18653/v1/2024.acl-long.818} {Same task, more tokens: the impact of input length on the reasoning performance of large language models}.
\newblock In \emph{Proceedings of the 62nd Annual Meeting of the Association for Computational Linguistics (Volume 1: Long Papers)}, pages 15339--15353, Bangkok, Thailand. Association for Computational Linguistics.

\bibitem[{Liu et~al.(2023)Liu, Yang, Shen, Hu, Zhang, Gu, and Zhang}]{liu2023think}
Lei Liu, Xiaoyan Yang, Yue Shen, Binbin Hu, Zhiqiang Zhang, Jinjie Gu, and Guannan Zhang. 2023.
\newblock \href {https://arxiv.org/abs/2311.08719} {Think-in-memory: Recalling and post-thinking enable llms with long-term memory}.
\newblock \emph{arXiv preprint arXiv:2311.08719}.

\bibitem[{Liu et~al.(2024{\natexlab{a}})Liu, Lin, Hewitt, Paranjape, Bevilacqua, Petroni, and Liang}]{lim}
Nelson~F. Liu, Kevin Lin, John Hewitt, Ashwin Paranjape, Michele Bevilacqua, Fabio Petroni, and Percy Liang. 2024{\natexlab{a}}.
\newblock \href {https://doi.org/10.1162/tacl_a_00638} {Lost in the middle: How language models use long contexts}.
\newblock \emph{Transactions of the Association for Computational Linguistics}, 12:157--173.

\bibitem[{Liu et~al.(2024{\natexlab{b}})Liu, Wang, and Chen}]{liu-etal-2024-rethinking}
Tianyang Liu, Fei Wang, and Muhao Chen. 2024{\natexlab{b}}.
\newblock \href {https://doi.org/10.18653/v1/2024.naacl-long.26} {Rethinking tabular data understanding with large language models}.
\newblock In \emph{Proceedings of the 2024 Conference of the North American Chapter of the Association for Computational Linguistics: Human Language Technologies (Volume 1: Long Papers)}, pages 450--482, Mexico City, Mexico. Association for Computational Linguistics.

\bibitem[{Liu et~al.(2025)Liu, Huang, Zeng, xinlong hao, Yu, Li, Wang, Gan, Liu, Yu, WANG, Wang, Ning, Hou, Wang, Wu, Xinzhi, Liu, Wang, Tang, Tu, Shang, Jiang, Tang, Lian, Liu, and Chen}]{liu2025toolace}
Weiwen Liu, Xu~Huang, Xingshan Zeng, xinlong hao, Shuai Yu, Dexun Li, Shuai Wang, Weinan Gan, Zhengying Liu, Yuanqing Yu, Zezhong WANG, Yuxian Wang, Wu~Ning, Yutai Hou, Bin Wang, Chuhan Wu, Wang Xinzhi, Yong Liu, Yasheng Wang, and 8 others. 2025.
\newblock \href {https://openreview.net/forum?id=8EB8k6DdCU} {Tool{ACE}: Winning the points of {LLM} function calling}.
\newblock In \emph{The Thirteenth International Conference on Learning Representations}.

\bibitem[{Marino(2025)}]{localiq}
Susie Marino. 2025.
\newblock \href {https://localiq.com/blog/chatbot-statistics/} {35+ chatbot statistics you need to know for 2025}.
\newblock LocaliQ.
\newblock Accessed: 29 Dec 2025.

\bibitem[{Meghwani et~al.(2025)Meghwani, Agarwal, Pattnayak, Patel, and Panda}]{meghwani2025hard}
Hansa Meghwani, Amit Agarwal, Priyaranjan Pattnayak, Hitesh~Laxmichand Patel, and Srikant Panda. 2025.
\newblock \href {https://doi.org/10.18653/v1/2025.acl-industry.72} {Hard negative mining for domain-specific retrieval in enterprise systems}.
\newblock In \emph{Proceedings of the 63rd Annual Meeting of the Association for Computational Linguistics (Volume 6: Industry Track)}, pages 1013--1026, Vienna, Austria. Association for Computational Linguistics.

\bibitem[{Mirzadeh et~al.(2025)Mirzadeh, Alizadeh, Shahrokhi, Tuzel, Bengio, and Farajtabar}]{mirzadeh2025gsmsymbolic}
Seyed~Iman Mirzadeh, Keivan Alizadeh, Hooman Shahrokhi, Oncel Tuzel, Samy Bengio, and Mehrdad Farajtabar. 2025.
\newblock \href {https://openreview.net/forum?id=AjXkRZIvjB} {{GSM}-symbolic: Understanding the limitations of mathematical reasoning in large language models}.
\newblock In \emph{The Thirteenth International Conference on Learning Representations}.

\bibitem[{Pan et~al.(2024)Pan, Wu, Jiang, Xia, Luo, Zhang, Lin, R{\"u}hle, Yang, Lin, Zhao, Qiu, and Zhang}]{pan-etal-2024-llmlingua}
Zhuoshi Pan, Qianhui Wu, Huiqiang Jiang, Menglin Xia, Xufang Luo, Jue Zhang, Qingwei Lin, Victor R{\"u}hle, Yuqing Yang, Chin-Yew Lin, H.~Vicky Zhao, Lili Qiu, and Dongmei Zhang. 2024.
\newblock \href {https://doi.org/10.18653/v1/2024.findings-acl.57} {{LLML}ingua-2: Data distillation for efficient and faithful task-agnostic prompt compression}.
\newblock In \emph{Findings of the Association for Computational Linguistics: ACL 2024}, pages 963--981, Bangkok, Thailand. Association for Computational Linguistics.

\bibitem[{Papineni et~al.(2002)Papineni, Roukos, Ward, and Zhu}]{papineni-etal-2002-bleu}
Kishore Papineni, Salim Roukos, Todd Ward, and Wei-Jing Zhu. 2002.
\newblock \href {https://doi.org/10.3115/1073083.1073135} {{B}leu: a method for automatic evaluation of machine translation}.
\newblock In \emph{Proceedings of the 40th Annual Meeting of the Association for Computational Linguistics}, pages 311--318, Philadelphia, Pennsylvania, USA. Association for Computational Linguistics.

\bibitem[{Parikh et~al.(2020)Parikh, Wang, Gehrmann, Faruqui, Dhingra, Yang, and Das}]{parikh-etal-2020-totto}
Ankur Parikh, Xuezhi Wang, Sebastian Gehrmann, Manaal Faruqui, Bhuwan Dhingra, Diyi Yang, and Dipanjan Das. 2020.
\newblock \href {https://doi.org/10.18653/v1/2020.emnlp-main.89} {{ToTTo}: A controlled table-to-text generation dataset}.
\newblock In \emph{Proceedings of the 2020 Conference on Empirical Methods in Natural Language Processing (EMNLP)}, pages 1173--1186, Online. Association for Computational Linguistics.

\bibitem[{Paruthi(2024)}]{glean}
Veraj Paruthi. 2024.
\newblock \href {https://www.glean.com/blog/glean-input-token-llm-latency} {How input token count impacts the latency of ai chat tools}.
\newblock Glean.
\newblock Accessed: 29 Dec 2025.

\bibitem[{Patil et~al.(2025{\natexlab{a}})Patil, Mao, Yan, Ji, Suresh, Stoica, and Gonzalez}]{patil2025bfcl}
Shishir~G Patil, Huanzhi Mao, Fanjia Yan, Charlie Cheng-Jie Ji, Vishnu Suresh, Ion Stoica, and Joseph~E. Gonzalez. 2025{\natexlab{a}}.
\newblock \href {https://proceedings.mlr.press/v267/patil25a.html} {The berkeley function calling leaderboard ({BFCL}): From tool use to agentic evaluation of large language models}.
\newblock In \emph{Proceedings of the 42nd International Conference on Machine Learning}, volume 267 of \emph{Proceedings of Machine Learning Research}, pages 48371--48392. PMLR.

\bibitem[{Patil et~al.(2025{\natexlab{b}})Patil, Mao, Yan, Ji, Suresh, Stoica, and Gonzalez}]{patil2025the}
Shishir~G Patil, Huanzhi Mao, Fanjia Yan, Charlie Cheng-Jie Ji, Vishnu Suresh, Ion Stoica, and Joseph~E. Gonzalez. 2025{\natexlab{b}}.
\newblock \href {https://openreview.net/forum?id=2GmDdhBdDk} {The berkeley function calling leaderboard ({BFCL}): From tool use to agentic evaluation of large language models}.
\newblock In \emph{Forty-second International Conference on Machine Learning}.

\bibitem[{Pattnayak et~al.(2025)Pattnayak, Agarwal, Meghwani, Patel, and Panda}]{pattnayak2025hybrid}
Priyaranjan Pattnayak, Amit Agarwal, Hansa Meghwani, Hitesh~Laxmichand Patel, and Srikant Panda. 2025.
\newblock \href {https://doi.org/10.18653/v1/2025.knowledgenlp-1.20} {Hybrid {AI} for responsive multi-turn online conversations with novel dynamic routing and feedback adaptation}.
\newblock In \emph{Proceedings of the 4th International Workshop on Knowledge-Augmented Methods for Natural Language Processing}, pages 215--229, Albuquerque, New Mexico, USA. Association for Computational Linguistics.

\bibitem[{Prasad et~al.(2024)Prasad, Koller, Hartmann, Clark, Sabharwal, Bansal, and Khot}]{prasad-etal-2024-adapt}
Archiki Prasad, Alexander Koller, Mareike Hartmann, Peter Clark, Ashish Sabharwal, Mohit Bansal, and Tushar Khot. 2024.
\newblock \href {https://doi.org/10.18653/v1/2024.findings-naacl.264} {{AD}a{PT}: As-needed decomposition and planning with language models}.
\newblock In \emph{Findings of the Association for Computational Linguistics: NAACL 2024}, pages 4226--4252, Mexico City, Mexico. Association for Computational Linguistics.

\bibitem[{Press et~al.(2022)Press, Smith, and Lewis}]{press2022tstl}
Ofir Press, Noah~A. Smith, and Mike Lewis. 2022.
\newblock \href {https://arxiv.org/abs/2108.12409} {Train short, test long: Attention with linear biases enables input length extrapolation}.
\newblock \emph{Preprint}, arXiv:2108.12409.

\bibitem[{Rafailov et~al.(2024)Rafailov, Sharma, Mitchell, Ermon, Manning, and Finn}]{rafailov2024dpo}
Rafael Rafailov, Archit Sharma, Eric Mitchell, Stefano Ermon, Christopher~D. Manning, and Chelsea Finn. 2024.
\newblock \href {https://arxiv.org/abs/2305.18290} {Direct preference optimization: Your language model is secretly a reward model}.
\newblock \emph{Preprint}, arXiv:2305.18290.

\bibitem[{Salama et~al.(2025)Salama, Cai, Yuan, Currey, Sunkara, Zhang, and Benajiba}]{salama-etal-2025-meminsight}
Rana Salama, Jason Cai, Michelle Yuan, Anna Currey, Monica Sunkara, Yi~Zhang, and Yassine Benajiba. 2025.
\newblock \href {https://doi.org/10.18653/v1/2025.emnlp-main.1683} {{M}em{I}nsight: Autonomous memory augmentation for {LLM} agents}.
\newblock In \emph{Proceedings of the 2025 Conference on Empirical Methods in Natural Language Processing}, pages 33136--33152, Suzhou, China. Association for Computational Linguistics.

\bibitem[{Sarangi et~al.(2025)Sarangi, Elgarf, and Salam}]{sarangi-etal-2025-decompose}
Sneheel Sarangi, Maha Elgarf, and Hanan Salam. 2025.
\newblock \href {https://aclanthology.org/2025.coling-main.682/} {Decompose-{T}o{M}: Enhancing theory of mind reasoning in large language models through simulation and task decomposition}.
\newblock In \emph{Proceedings of the 31st International Conference on Computational Linguistics}, pages 10228--10241, Abu Dhabi, UAE. Association for Computational Linguistics.

\bibitem[{Singh(2021)}]{Singh2021}
J.~Singh. 2021.
\newblock \href {https://doi.org/10.25080/majora-1b6fd038-009} {Social media analysis using natural language processing techniques}.
\newblock In \emph{Proceedings of the 20th Python in Science Conference}, SciPy, page 74–80. SciPy.

\bibitem[{Singh(2022)}]{Singh2022}
Jyotika Singh. 2022.
\newblock \href {https://doi.org/10.25080/majora-212e5952-017} {pyaudioprocessing: Audio processing, feature extraction, and machine learning modeling}.
\newblock In \emph{Proceedings of the 21st Python in Science Conference}, SciPy, page 152–158. SciPy.

\bibitem[{Singh(2023)}]{Singh2023}
Jyotika Singh. 2023.
\newblock \href {https://doi.org/10.1201/9781003264774} {\emph{Natural Language Processing in the Real World: Text Processing, Analytics, and Classification}}.
\newblock Chapman and Hall/CRC.

\bibitem[{Singh et~al.(2021)Singh, Bilbro, Avon, Bowen, Jolicoeur, and Matta}]{icxpatent}
Jyotika Singh, Rebecca Bilbro, Michael Avon, Scott Bowen, Dan Jolicoeur, and Serge Matta. 2021.
\newblock \href {https://patents.google.com/patent/US10949880B2/} {Method for optimizing media and marketing content using cross-platform video intelligence}.
\newblock U.S. Patent.
\newblock Issued March 16, 2021.

\bibitem[{Singh et~al.(2025)Singh, Sun, Agarwal, Krishnamurthy, Benajiba, Ravi, and Roth}]{singh-etal-2025-llms}
Jyotika Singh, Weiyi Sun, Amit Agarwal, Viji Krishnamurthy, Yassine Benajiba, Sujith Ravi, and Dan Roth. 2025.
\newblock \href {https://doi.org/10.18653/v1/2025.emnlp-industry.60} {Can {LLM}s narrate tabular data? an evaluation framework for natural language representations of text-to-{SQL} system outputs}.
\newblock In \emph{Proceedings of the 2025 Conference on Empirical Methods in Natural Language Processing: Industry Track}, pages 883--902, Suzhou (China). Association for Computational Linguistics.

\bibitem[{Su et~al.(2023)Su, Lu, Pan, Murtadha, Wen, and Liu}]{su2023rope}
Jianlin Su, Yu~Lu, Shengfeng Pan, Ahmed Murtadha, Bo~Wen, and Yunfeng Liu. 2023.
\newblock \href {https://arxiv.org/abs/2104.09864} {Roformer: Enhanced transformer with rotary position embedding}.
\newblock \emph{Preprint}, arXiv:2104.09864.

\bibitem[{team(2025)}]{blockgit}
Goose team. 2025.
\newblock \href {https://block.github.io/goose/docs/guides/sessions/smart-context-management} {Smart context management}.
\newblock Goose.
\newblock Accessed: 29 Dec 2025.

\bibitem[{Vaswani et~al.(2017)Vaswani, Shazeer, Parmar, Uszkoreit, Jones, Gomez, Kaiser, and Polosukhin}]{10.5555/3295222.3295349}
Ashish Vaswani, Noam Shazeer, Niki Parmar, Jakob Uszkoreit, Llion Jones, Aidan~N. Gomez, \L{}ukasz Kaiser, and Illia Polosukhin. 2017.
\newblock Attention is all you need.
\newblock In \emph{Proceedings of the 31st International Conference on Neural Information Processing Systems}, NIPS'17, page 6000–6010, Red Hook, NY, USA. Curran Associates Inc.

\bibitem[{Wang et~al.(2025{\natexlab{a}})Wang, Fu, Cao, Wang, Tian, and Ding}]{wang2025recursively}
Qingyue Wang, Yanhe Fu, Yanan Cao, Shuai Wang, Zhiliang Tian, and Liang Ding. 2025{\natexlab{a}}.
\newblock \href {https://doi.org/10.1016/j.neucom.2025.130193} {Recursively summarizing enables long-term dialogue memory in large language models}.
\newblock \emph{Neurocomput.}, 639(C).

\bibitem[{Wang et~al.(2025{\natexlab{b}})Wang, Wu, Yao, and Su}]{WANG2025107200}
Yaoxiang Wang, Zhiyong Wu, Junfeng Yao, and Jinsong Su. 2025{\natexlab{b}}.
\newblock \href {https://doi.org/10.1016/j.neunet.2025.107200} {Tdag: A multi-agent framework based on dynamic task decomposition and agent generation}.
\newblock \emph{Neural Networks}, 185:107200.

\bibitem[{Wen et~al.(2024)Wen, Zhong, Ke, Shao, Wang, and Huang}]{wen-etal-2024-learning}
Jiaxin Wen, Ruiqi Zhong, Pei Ke, Zhihong Shao, Hongning Wang, and Minlie Huang. 2024.
\newblock \href {https://doi.org/10.18653/v1/2024.acl-long.629} {Learning task decomposition to assist humans in competitive programming}.
\newblock In \emph{Proceedings of the 62nd Annual Meeting of the Association for Computational Linguistics (Volume 1: Long Papers)}, pages 11700--11723, Bangkok, Thailand. Association for Computational Linguistics.

\bibitem[{Xu et~al.(2022)Xu, Szlam, and Weston}]{xu2022beyond}
Jing Xu, Arthur Szlam, and Jason Weston. 2022.
\newblock \href {https://doi.org/10.18653/v1/2022.acl-long.356} {Beyond goldfish memory: Long-term open-domain conversation}.
\newblock In \emph{Proceedings of the 60th Annual Meeting of the Association for Computational Linguistics (Volume 1: Long Papers)}, pages 5180--5197, Dublin, Ireland. Association for Computational Linguistics.

\bibitem[{Yang et~al.(2025)Yang, Some, Bain, and Kang}]{YANG2025113503}
Wenli Yang, Lilian Some, Michael Bain, and Byeong Kang. 2025.
\newblock \href {https://doi.org/10.1016/j.knosys.2025.113503} {A comprehensive survey on integrating large language models with knowledge-based methods}.
\newblock \emph{Knowledge-Based Systems}, 318:113503.

\bibitem[{Yu et~al.(2018)Yu, Zhang, Yang, Yasunaga, Wang, Li, Ma, Li, Yao, Roman, Zhang, and Radev}]{yu-etal-2018-spider}
Tao Yu, Rui Zhang, Kai Yang, Michihiro Yasunaga, Dongxu Wang, Zifan Li, James Ma, Irene Li, Qingning Yao, Shanelle Roman, Zilin Zhang, and Dragomir Radev. 2018.
\newblock \href {https://doi.org/10.18653/v1/D18-1425} {{S}pider: A large-scale human-labeled dataset for complex and cross-domain semantic parsing and text-to-{SQL} task}.
\newblock In \emph{Proceedings of the 2018 Conference on Empirical Methods in Natural Language Processing}, pages 3911--3921, Brussels, Belgium. Association for Computational Linguistics.

\bibitem[{Yuan et~al.(2026)Yuan, Sun, Rezaeian, Singh, Ghoshal, Wang, Ballesteros, and Benajiba}]{yuan2026barriersdiscretereasoningtransformers}
Michelle Yuan, Weiyi Sun, Amir~H. Rezaeian, Jyotika Singh, Sandip Ghoshal, Yao-Ting Wang, Miguel Ballesteros, and Yassine Benajiba. 2026.
\newblock \href {https://arxiv.org/abs/2602.11175} {Barriers to discrete reasoning with transformers: A survey across depth, exactness, and bandwidth}.
\newblock \emph{Preprint}, arXiv:2602.11175.

\bibitem[{Zhang et~al.(2025)Zhang, Dai, Wu, Yang, Wang, Tang, and Liu}]{zhang2025survey}
Chen Zhang, Xinyi Dai, Yaxiong Wu, Qu~Yang, Yasheng Wang, Ruiming Tang, and Yong Liu. 2025.
\newblock \href {https://arxiv.org/abs/2501.09959} {A survey on multi-turn interaction capabilities of large language models}.
\newblock \emph{arXiv preprint arXiv:2501.09959}.

\bibitem[{Zhang et~al.(2022)Zhang, Liu, Li, Zeng, Wang, You, Miao, and Cui}]{zhang2022history}
Tong Zhang, Yong Liu, Boyang Li, Zhiwei Zeng, Pengwei Wang, Yuan You, Chunyan Miao, and Lizhen Cui. 2022.
\newblock \href {https://doi.org/10.18653/v1/2022.findings-emnlp.247} {History-aware hierarchical transformer for multi-session open-domain dialogue system}.
\newblock In \emph{Findings of the Association for Computational Linguistics: EMNLP 2022}, pages 3395--3407, Abu Dhabi, United Arab Emirates. Association for Computational Linguistics.

\bibitem[{Zheng et~al.(2024)Zheng, Chiang, Sheng, Li, Zhuang, Wu, Zhuang, Li, Lin, Xing, Gonzalez, Stoica, and Zhang}]{zheng2024lmsyschatm}
Lianmin Zheng, Wei-Lin Chiang, Ying Sheng, Tianle Li, Siyuan Zhuang, Zhanghao Wu, Yonghao Zhuang, Zhuohan Li, Zi~Lin, Eric Xing, Joseph~E. Gonzalez, Ion Stoica, and Hao Zhang. 2024.
\newblock \href {https://openreview.net/forum?id=BOfDKxfwt0} {{LMSYS}-chat-1m: A large-scale real-world {LLM} conversation dataset}.
\newblock In \emph{The Twelfth International Conference on Learning Representations}.

\end{thebibliography}

\newpage
\appendix
\section{Extended related work}
\label{sec:extrw}

Chat systems in Natural Language Processing (NLP) (field comprising text and audio processing) have existed well before large language models (LLMs) became state of the art~\cite{10.5555/3295222.3295349, Singh2022}. NLP applications are widely used across industry verticals~\cite{Singh2023} and diverse data sources, including business and finance, social media, and even unstructured data~\cite{YANG2025113503, singh-etal-2025-llms, liu-etal-2024-rethinking,Singh2021}, spanning task-oriented, data-analysis-focused~\cite{icxpatent}, and free-form settings, and now LLMs are heavily used across these data types and applications.
While LLMs offer substantial advantages along with some limitations~\citep{yuan2026barriersdiscretereasoningtransformers}, they have made real-world chat applications far more prevalent by reducing reliance on separate intent/entity detection models and hand-crafted rules to control behavior based on conversational signals. Still, reliance on single-turn benchmarks prevail and multi-turn chats underperform single-turn specified prompts.

\paragraph{Memory compression}
\citet{chen2025compress} introduce the idea of compressive memory where they integrate conversation summaries with information about the user. They use Supervised Fine-Tuning (SFT) and Direct Preference Optimization (DPO) to train a Llama-2 model to summarize and compress memories without dependence on external modules. \citet{jiang-etal-2024-longllmlingua} uses a small language model to calculate the perplexity of each token in the original prompt and then remove tokens with lower perplexities, and reorders relevant parts to the beginning and end to combat the "lost in the middle" problem for long-context scenarios. It requires re-compression for different questions, even with the same context, preventing caching of the context. In addition, aggressive compression can inadvertently remove key details, impacting accuracy for tasks requiring nuance. 

\paragraph{Why long-context work is not competitive but complimentary}
Long-context compression methods such as LLMLingua-2~\cite{pan-etal-2024-llmlingua} and LongLLMLingua~\cite{jiang-etal-2024-longllmlingua} target a different regime of compressing very long inputs via token/segment pruning (often importance/perplexity-based) and, for LongLLMLingua, reordering/filtering. In short-horizon multi-turn setting, redundancy is limited and correctness often hinges on small but decisive state (e.g., negations, corrections like 'instead,' key entity mentions), so token-level deletion can disproportionately change intent or break coreference; reordering is also less meaningful when history is short and chronologically causal. Therefore, they are not an apples-to-apples comparison and not a suitable baseline for MT-OSC; rather, they are best viewed as prompt-compression tools that may be used in addition to MT-OSC for very long histories.

\paragraph{Multi-turn sharded datasets and hinted solution limitation}
\citet{laban2025llms} formulates a unique evaluation method for underspecification in multi-turn conversations. They introduce a sharding process where they transform datasets with fully-specified instructions into sharded instructions that are revealed on subsequent turns. 
The CONCAT method proposed in the paper could perform better in underspecified settings but is impractical for real-world scenarios, as users are unlikely to resend all prior context with each message. If automated, it is also ineffective for episodic multi-turn conversations, where the model may struggle to distinguish between the current and previous tasks. Additionally, the approach significantly increases the number of tokens in the conversation, reducing overall efficiency. There are two other methods mentioned in the paper, RECAP and SNOWBALL. Although RECAP performs better than SNOWBALL, its intervention is conducted on the last turn of the conversation, which is not known a priori, and both have been shown to have worse performance than CONCAT.
The two recommendations in the paper to mitigate information loss, "If time allows, try again." and "Consolidate before retrying.", are not principled approaches as they both require human evaluation and intervention. 

\section{Method prompts and details}
\label{sec:prompts}
\subsection{Condenser prompt}
The prompt for \textit{Condenser} is as follows. The examples used are shared in Figure~\ref{fig:fewshots}.

The inference parameter settings were $temperature=0.01$, $frequency\_penalty=1$, $max\_completion\_tokens=10000$, and $top\_p=1$, and LLM was \textsc{Llama-3.3-70B-Instruct}.

\noindent\fbox{%
    \begin{minipage}{0.48\textwidth} 
Condense the information from HumanInput and also share a concise summary of Assistant response to the human input. Make sure you don't miss any specific values and instructions provided by the human input that are relevant for the conversation.
For the assistant response summary, make sure to keep any important points but keep it concise. \\
Also return the reasoning behind your condensation strategy.
Return in JSON format  {{"HumanInput": "<text>", "Assistant": "<text>", "Reasoning": "<text>"}}. Only return the JSON with no additional text. \\

Examples: \\

Conversation History: <example-1>\\
<assistant-output-example-1>\\

Conversation History: <example-2>\\
<assistant-output-example-2>\\

Conversation History: <example-3>\\
<assistant-output-example-3>\\

Conversation History: 
    \end{minipage}%
}

\subsection{Simple summarization prompt}
The prompt for the simple summarization is as follows.

\noindent\fbox{%
    \begin{minipage}{0.48\textwidth} 
Summarize the information from HumanInput and also share a concise summary of Assistant response to the human input. Make sure you don't miss any specific values and instructions provided by the human input. For the assistant response summary, make sure to keep any important points but keep it concise. Return in JSON format  {{"HumanInput": "<text>", "Assistant": "<text>"}}. Only return the JSON with no additional text.
    \end{minipage}%
}

The inference parameter settings were $temperature=0.01$, $frequency\_penalty=1$, $max\_completion\_tokens=10000$, and $top\_p=1$, and LLM was \textsc{Llama-3.3-70B-Instruct}.

\subsection{MT-OSC details}
For the Decider, we used $\gamma$ 0.2 and $\tau$ 1000 tokens.

The one-off sequential condenser method (MT-OSC) for w=4 can be represented as follows in equation form.

\textbf{Definitions:}
\begin{itemize}
    \item $u_i$: User message at turn $i$
    \item $a_i$: Assistant message at turn $i$
    \item $C_j = (C_{ju}, C_{ja})$: Output of the $j$-th condenser, summarized user-assistant pair
    \item $w$: Window size (number of user-assistant pairs condensed at a time)
\end{itemize}

\textbf{Chat History Representation:}
\begin{align*}
    &\text{At turn 1:} \quad H_1 = \emptyset\\
    &\text{At turn 2:} \quad H_2 = \{ (u_1, a_1) \}\\
    &\text{At turn 3:} \quad H_3 = \{ (u_1,a_1), (u_2,a_2) \}\\
    &\text{At turn 4:} \quad H_4 = \{ (u_1,a_1), (u_2,a_2), (u_3,a_3) \}
\end{align*}

When $|H_t| = w$ (e.g., $w=4$), condensation is triggered as a background process:
\begin{align*}
C_1 =\ & \text{Condense}\left( \{ (u_1, a_1),\ (u_2, a_2), \right. \\
      & \left. (u_3, a_3),\ (u_4, a_4) \} \right ) = (C_{1u}, C_{1a})
\end{align*}

\textbf{Turn 5 continues while $C_1$ is generating:}
\begin{align*}
&\text{At turn 5:} \quad H_5 = \{ (u_1,a_1), (u_2,a_2), (u_3,a_3), \\
& \qquad \qquad \qquad \qquad u_4,a_4) \}
\end{align*}

\textbf{After $C_1$ is generated:}
\begin{align*}
    &\text{At turn 6:} \quad H_6 = \{ (C_{1u}, C_{1a}), (u_5,a_5) \}\\
    &\text{At turn 7:} \quad H_7 = \{ (C_{1u}, C_{1a}), (u_5,a_5),
    (u_6,a_6)\\
\end{align*}

When $H_t$ contains $w$ new pairs, condensation is triggered again:

\begin{multline*}
C_2 = \text{Condense}\Big( \{\, (C_{1u}, C_{1a}),\ (u_5, a_5), \\
            (u_6, a_6)\, (u_7, a_7)\, \} \Big) = (C_{2u}, C_{2a})
\end{multline*}

\textbf{Turn 8 continues while $C_2$ is generating:}
\noindent\begin{multline*}
\text{At turn 8:} \quad H_8 = \{ (C_{1u}, C_{1a}),\ (u_5,a_5), \\
\qquad (u_6,a_6),\ (u_7,a_7) \}
\end{multline*}

\textbf{After $C_2$ is generated:}
\noindent\begin{multline*}
\text{At turn 9:} \quad H_9 = \{ (C_{2u}, C_{2a}),\ (u_8,a_8) \}\\
\qquad\text{and~so~on}
\end{multline*}

\section{Dataset details}
\label{sec:datasetext}
In general, there has been work done to study conversations across LLMs and applications \cite{pattnayak2025hybrid, meghwani2025hard, agarwal2025aligning}, but not specific to multi-turn short-term memory. Below, we dig into this problem and closely related datasets specific to our paper's multi-turn conversation scope in single chat sessions.

\begin{figure}
    \centering
    \includegraphics[width=1\linewidth]{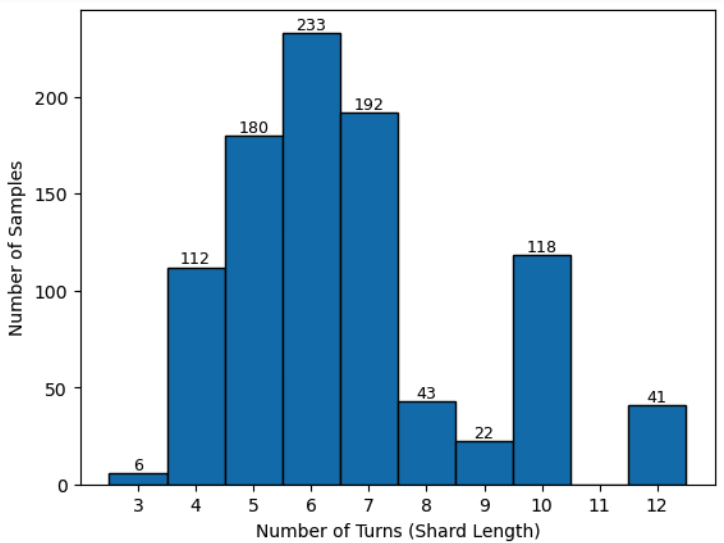}
    \caption{Histogram showing the number of samples for each maximum number of turns per sample in sharded + MTEval datasets.}
    \label{fig:turnsample}
\end{figure}

\paragraph{Datasets choice and details:}  While other multi-turn datasets exist in the literature, many contain only short conversations—for example, ToolAce \cite{liu2025toolace} is limited to a maximum of four turns, and the more recent BFCL multi-turn and multi-step function calling dataset \cite{patil2025bfcl} has very few examples exceeding five turns. The datasets we selected for our experiments contained up to 12 turns. Figure \ref{fig:turnsample} shows the number of samples by number of turns for the datasets used for our experiments (sharded and MTEval datasets combined).

\begin{table*}[h!]
\centering
\begin{tabular}{|p{3.5cm}|p{2cm}|p{4cm}|c|p{3.3cm}|}
\hline
Dataset  & Source & Success Metric & Size & Split (Dev, Val, Test)\\
\hline
Sharded BFCL & \cite{laban2025llms} & Accuracy [Exact match (Action matching)]
& 105 & (30.5\%, 19\%, 50.5\%)\\
Sharded GSM8K & \cite{laban2025llms} & Accuracy [Exact match (Final Answer)]
& 103 & (-, 36\%, 64\%)\\ 
Sharded HumanEval  & \cite{laban2025llms} & Accuracy [Exact match (Code test passing)] & 45 & (-, -, 100\%) \\
Sharded Spider  & \cite{laban2025llms} & Accuracy [Exact match (SQL execution output match)] & 107 & (-, -, 100\%) \\
Sharded ToTTo  & \cite{laban2025llms} & BLEU Score & 120 & (15\%, 12.5\%, 72.5\%) \\
Sharded Summary of Haystack  & \cite{laban2025llms} & LLMaaJ ("Joint Score") to measure info coverage \& attribution F1 & 94 & (-, -, 100\%) \\
MT-EVAL refinement & \cite{kwan2024mt} & LLMaaJ for quality rating out of 10 & 40 & (-, -, 100\%)\\
MT-EVAL follow-up  & \cite{kwan2024mt} & LLMaaJ for quality rating out of 10 & 60 & (-, -, 100\%) \\
MT-EVAL recollection+  & \cite{kwan2024mt} & Accuracy [Exact match (Final Answer)] & 100 & (-, -, 100\%)\\
MT-EVAL expansion+  & \cite{kwan2024mt} & LLMaaJ for quality rating out of 10 & 100 & (-, -, 100\%)\\
\hline
\end{tabular}
\caption{Dataset name, source, success metric for evaluation, size, and split. The sharded datasets were synthetically generated and manual inspection and validation by the authors of their respective sources cited above. The MT-EVAL datasets also went through human-in-the-loop creation, validation, and filtering process where the initial data generation utilized GPT-4.}
\label{tab:dataset}
\end{table*}

Table ~\ref{tab:dataset} shows a breakdown of datasets used, their source, sizes, splits between development, validation, and test sets. All samples in the dataset are less than or equal to 12 turns, with a vast majority of them being <=10 turns.
The total size of our development set is 50 samples, validation set is 72 samples, and test set (on which we report all the results in the paper) is 752 samples.

The development set was primarily used for manual inspection and for analyzing the impact of different experimental techniques and prompt variations. As a result, the development subset is smaller in size relative to the validation and test subsets. To ensure unbiased evaluation, we strictly excluded any overlap between the datasets used in the development phase and those used in validation and testing. In particular, all four MT-EVAL datasets were reserved exclusively for the test set. Among the six sharded datasets, some were allocated entirely to the test and/or validation sets, while others were divided across the development, validation, and test sets.

\subsection{Evaluation details}
\label{sec:evalappend}
Evaluation metrics per dataset are shared in Table~\ref{tab:dataset}.

For all datasets, we adhere as closely as possible to the original evaluation methods, making only minimal necessary assumptions. There has been more work on evaluation of multi-turn conversations using LLM-as-a-Judge \cite{duan-etal-2024-botchat}, however those are free-form chats rather as task-oriented and hence not applicable to our work.

Sharded versions~\cite{laban2025llms} of HumanEval~\cite{Chen2021EvaluatingLL}, Spider~\cite{yu-etal-2018-spider}, BFCL-V3 Parallel~\cite{patil2025the}, and GSM8K~\cite{Cobbe2021TrainingVT} are assessed based on binary correctness—either through execution (code in HumanEval and SQL queries in Spider) or by checking semantic equivalence to a reference answer (API calls in BFCL and numerical responses in GSM). Each sample consists of multiple conversational turns, with each final answer attempt evaluated for correctness. If a correct answer is produced at any point during the conversation, the sample is considered a success.

Sharded versions of ToTTo~\cite{parikh-etal-2020-totto} and Summary of Haystack~\cite{laban-etal-2024-summary} datasets resemble refinement tasks evaluated on a continuous scale from 0 to 1. ToTTo uses the BLEU metric~\cite{papineni-etal-2002-bleu}, while Summary of Haystack is scored using a custom LLM-as-a-judge metric called Joint Score, designed to assess the summary’s information coverage and attribution accuracy. For both tasks, we evaluate the model’s response at the final turn of the conversation, as it is expected that earlier turns will progressively refine the answer leading up to this point for these datasets. 

MT-Eval datasets (inspired by LMSYS-Chat-1M ~\cite{zheng2024lmsyschatm} and MT-Bench~\cite{bai-etal-2024-mt}) for Refinement, Expansion, and Follow-up are passed through an LLM judge to rate answer for each turn in a sample, which is then averaged to get an average score per sample in the dataset.  MT-Eval Recollection dataset uses exact match for category output by the model based on ground-truth categories.

\paragraph{Sharded GSM8K evaluation:} Our evaluation approach for sharded GSM8K  dataset diverges from the rule-based method used in the source paper, which we found is prone to false negatives—particularly when the numerical answer is not positioned at the end of the response string. For example, the method marks “for the 5 items, the price is \$78” as correct, but “the price is \$78 for the 5 items” as incorrect, despite their semantic equivalence. Manual inspection of mismatches revealed multiple cases where identical or correct answers failed to match due to these limitations. For our final reported results, we passed the responses not matched by the rule-based methods to be subsequently assessed by an LLM-as-a-Judge using \textsc{Llama-3.3-70B-Instruct} with $temperature=0$, $top\_p=1$, and $max\_tokens=4096$. Manual inspection revealed resolution of false negatives using this 2-tier evaluation approach for this dataset. The judge prompt used is as follows:
\\
\noindent\fbox{%
    \begin{minipage}{0.48\textwidth} 

I am sharing a user Q and a reference answer for that question. Then I will show you a model generated answer.\\
Return True if the model generated answer is correct and the final answer it contains matches the answer in the reference text. Otherwise, return False.\\

User Q: \{\}\\

Reference answer: \{\}\\

Model generated answer: \{\} 
\end{minipage}}

\subsection{MT-EVAL Synthesis}
\label{sec:mtevalsyn}

Two out of the four MT-EVAL datasets, namely \textbf{Recollection} and \textbf{Expansion}, originally contained only 10 samples each. To increase this number to 100 samples per dataset, we used the following methods:

\paragraph{MT-EVAL Recollection Dataset}
Each sample consists of a user instruction followed by document text in all subsequent turns. The dataset contains 100 unique documents in total, with 10 documents (one per turn) used in each sample. To generate additional samples, we randomly selected 10 different documents from the pool of 100 for each new sample, while keeping the top-level user instruction and task unchanged.

\paragraph{MT-EVAL Expansion Dataset}
Each sample begins with a document in the first turn, followed by 10 different questions related to that document in subsequent turns. These questions depend only on the initial document and are not order-sensitive. To create variants, we shuffled the order of the 10 questions for each sample, producing 9 new variants per original sample. This process, combined with the 10 original samples, resulted in a total of 100 samples for this dataset.

\subsection{Generating distracting turns}
\label{sec:promptnoop}

We observe that most publicly available multi-turn datasets are relatively clean with no distractions or noise. To further assess the robustness of our experiments, we introduced simulated redundancy and task-irrelevant information between turns within the above datasets. This modification allowed us to evaluate performance in less curated, more realistic conversational settings, thereby ensuring our results generalize beyond cleaner, idealized data. Specifically, we introduced the following types of redundancy and irrelevant information into the multi-turn conversations:

\begin{enumerate}
\item \textbf{Repetition Infusion}: Repeated random turns within the multi-turn dialogues -  for each multi-turn conversation, we randomly select N turns in the middle (not the first turn or the last turn) and repeat them once in the same position. N is set to be a portion of the total number of turns in the conversation.
\item \textbf{Filler Injection}: Inserted random turns containing only meaningless filler words, such as 'Um.', 'Uh.', 'Well.', 'Anyway.', 'Ok.', 'Hmm.', etc. - for each multi-turn conversation, we randomly select N turns in the middle (exclude the first turn) and add a randomly selected filler word as extra turn right before them.
\item \textbf{Contextual Diversion}: 
For each multi-turn chat, we randomly select N turns from the middle (excluding the first and last) and add new, contextually relevant turns that aren’t essential for task completion. If a selected turn isn’t blank, we generate an extra turn based on its content and place it immediately after.
Appendix~\ref{sec:promptnoop} shows the prompt we use to generate the extra distracting turns.
\end{enumerate}

We used the model \textsc{GPT4o1-200B-128} to generate the distracting turns for Contextual Diversion with inference parameter settings as follows: $temperature = 1.0$, $top\_p = 0.75$, $max\_tokens = 4096$.

\noindent\fbox{%
    \begin{minipage}{0.48\textwidth} 
Can you come up with related but distracting information as new entry in user inputs, for example:\\
\\
Input:\\
Check for availability of meeting rooms for next week.\\
\\
Output:\\
Meeting rooms are often in high demand, especially during peak business hours. It's a good idea to consider alternative locations or virtual meeting options in case the preferred rooms are fully booked. Additionally, checking the amenities available in each room, such as video conferencing equipment or whiteboards, can help ensure that the meeting runs smoothly.\\
\\
Make sure the added information is not helping make the instructions more clear.\\
\\
Please respond only with the output without any extra explanations or text.\\
\\
Here is the input to modify:
    \end{minipage}%
}

\subsubsection{Samples for Robustness Tests}
\label{sec:perturbedsample}

\begin{figure*}
    \centering
    \includegraphics[width=1\linewidth]{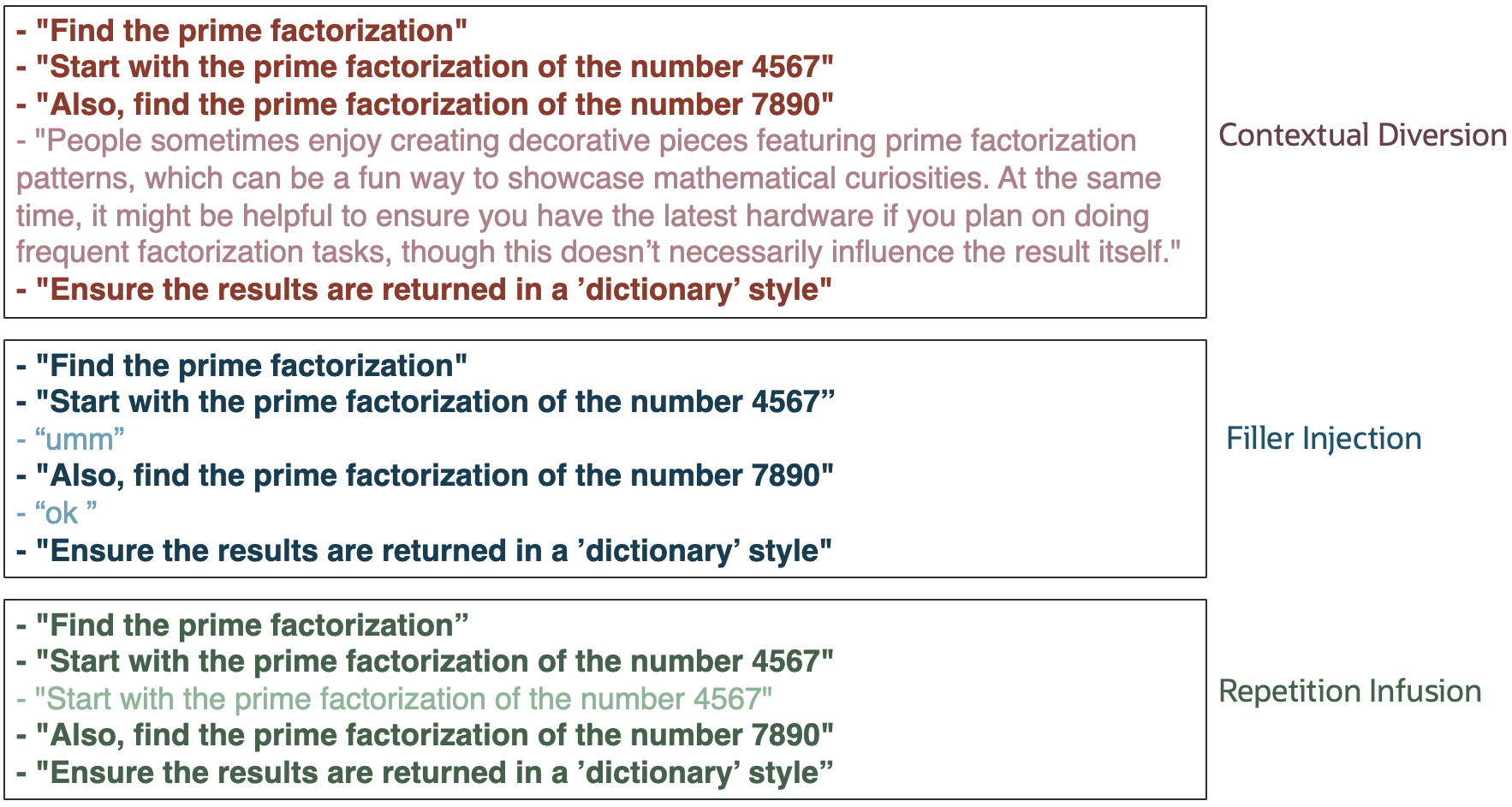}
    \caption{These examples shows the original turns from a conversation (in darker color in each example) and the perturbed version with new random turns (shown in the lighter color in each example) added to the conversation.}
    \label{fig:robdata}
\end{figure*}

Figure~\ref{fig:robdata} contains examples for data samples and how they were augmented for robustness tests for \emph{Repetition Infusion}, \emph{Filler Injection}, and \emph{Contextual Diversion}. The light color turns show the augmentations over existing data.

\section{Results}
\label{sec:resultsext}

\begin{figure*}
    \centering
    \includegraphics[width=1\linewidth]{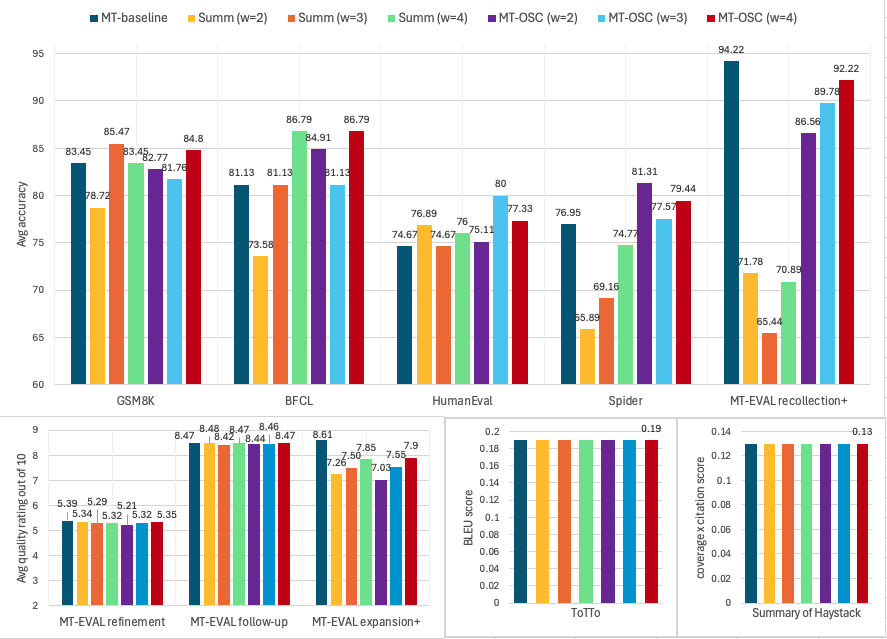}
    \caption{Comparison of performance across datasets of simple summarization (Summ) using the one-off sequential application approach (window sizes 2, 3, and 4) and the MT-OSC (window sizes 2, 3, and 4). Chat model is Llama-3.3-70B-Instruct.}
    \label{fig:resind}
\end{figure*}

Across inferences against different models used for analysis, the parameters were set to the following settings: $temperature=0.01$, $frequency\_penalty = 1$ (for LLMs that support this inference parameter), $max\_completion\_tokens=10000$, and $top\_p=1$.

Figure \ref{fig:resind} shows performance across of datasets using raw chat turns in model's context (MT (baseline)), using one-off sequential approach using simple summarizer prompt, and using MT-OSC. Results are across window sizes 2, 3, and 4. The results show largely maintained or improved performance with MT-OSC. 

\paragraph{FIFO}
We also experimented with a first-in-first-out (FIFO) context strategy; however, in our task-oriented datasets, relevant information often appears throughout early turns, so FIFO dropped critical context and resulted in substantially lower accuracy, and we therefore omit it from reporting in our extensive evaluations. Table~\ref{tab:fifo} shows FIFO results on our Sharded BFCL test set compared to MT-baseline and MT-OSC for chat models Llama-3.3-70B and GPT 4.1. Since it underperforms the industry-standard default MT-baseline, we do not include it in the main experiments. We set the FIFO limit to 4 turns: when the chat history exceeds 4 turns, the earliest turn is discarded. This matches MT-OSC with window size 4, where the chat history for any given turn contains at most four prior turns.

\begin{table}[t]
\centering
\begin{tabular}{lcc}
\hline
 & Llama-3.3-70B & GPT 4.1 \\
\hline
\fbox{\textbf{MT-baseline}} & 81.13\% & 61.54\% \\
\textbf{MT-OSC*}      & 86.79\% & 69.23\% \\
\textbf{FIFO} & \textbf{80.77\% $\downarrow$} & \textbf{51.92\% $\downarrow$} \\
\hline
\end{tabular}
\caption{FIFO compared to MT-baseline and MT-OSC for managing chat history shows that FIFO performs worse than the baseline (MT-baseline) approach on the Sharded BFCL dataset. *=ours}
\label{tab:fifo}
\end{table}

\subsection{Results across 13 LLMs}
\label{sec:resllms}

Table~\ref{tab:llmsdatasetsres} shows results for datasets across 13 LLMs for MT-baseline and MT-OSC.

\begin{table*}[h!]
\centering
\begin{tabular}{r|c|c|c|c|c|c|c||l}
\toprule
        & GSM & BFCL & HEval & Recoll & Spider & Expan & Refin & Avg\\
        & $^{shrd}$ & $^{shrd}$ & $^{shrd}$ & $^{mte+}$ & $^{shrd}$* & $^{mte+}$  & $^{mte}$ & $^{acc\%}$\\
\hline
Metric & \multicolumn{5}{c|}{accuracy \%} & \multicolumn{2}{c||}{rating out of 10} &\\
\hline
\multicolumn{5}{l}{\textbf{GROK3}} \\   
\texttt{MT} & 0.82     & 0.67    & 0.51         & 0.96 & 0.5 & 9.01 & 6.26 & 0.75\\
\texttt{MTOSC} & 0.86     & 0.71    & 0.62         & 0.95 &0.5 & 8.38 & 6.22 & \textbf{0.79}\\
\multicolumn{5}{l}{\textbf{GROK4}} \\   
\texttt{MT} & 0.84     & 0.64    & 0.84         & 0.80 & 0.5 & 9.05 & 5.92 & 0.78\\
\texttt{MTOSC}     & 0.86     & 0.69    & 0.80         & 0.86 &0.83 & 8.37 & 6.19 & \textbf{0.81}\\
\multicolumn{5}{l}{\textbf{Llama-3.1-405B}} \\   
\texttt{MT} & 0.82     & 0.82    & 0.69         & 0.92 &0.67 & 8.65 & 5.52 & 0.81\\
\texttt{MTOSC}  & 0.86    & 0.84    & 0.74         & 0.92  &0.5 & 7.98 & 5.51 & \textbf{0.84}\\
\multicolumn{5}{l}{\textbf{Llama-3.3-70B}} \\   
\texttt{MT} & 0.8     & 0.84    & 0.80         & 0.94 &0.5 & 8.61 & 5.39 & 0.84\\
\texttt{MTOSC}   & 0.8    & 0.87    & 0.84         & 0.92 &0.88  & 7.9 & 5.18 & \textbf{0.86}\\
\multicolumn{5}{l}{\textbf{Llama-4-Maverick}} \\   
\texttt{MT} & 0.76     & 0.84    & 0.83         & 0.90 &0.5 & 8.58 & 5.36 & 0.85\\
\texttt{MTOSC}   & 0.78     & 0.90    & 0.91         & 0.91 &0.67 & 7.82 & 5.43 & \textbf{0.88}\\
\multicolumn{5}{l}{\textbf{Llama-4-Scout}} \\   
\texttt{MT} & 0.84     & 0.88    & 0.67         & 0.92 &0.17 & 8.48 & 5.22 & 0.83\\
\texttt{MTOSC}  & 0.86     & 0.88    & 0.70        & 0.91 &0.83 & 7.65 & 5.34 & \textbf{0.84}\\
\multicolumn{5}{l}{\textbf{GPT 4.1}} \\   
\texttt{MT} & 0.92     & 0.67    & 0.83         & 0.97 &0.83 & 9.08 & 6.15 & 0.86\\
\texttt{MTOSC}    & 0.87     & 0.80    & 0.90         & 0.96 &0.67  & 8.41 & 6.26 & \textbf{0.89}\\
\multicolumn{5}{l}{\textbf{GPT 4.1 mini}} \\   
\texttt{MT} & 0.89     & 0.67    & 0.86         & 0.97 &0.5 & 8.97 & 6.04 & 0.85\\
\texttt{MTOSC}  & 0.83    & 0.79    & 0.90         & 0.96   &0.5 & 8.32 &  6.07 & \textbf{0.88}\\
\multicolumn{5}{l}{\textbf{GPT 4.1 nano}} \\   
\texttt{MT} & 0.70     & 0.51    & 0.70         & 0.86 &0.67 & 8.14 & 5.90 & 0.70\\
\texttt{MTOSC}   & 0.77     & 0.59    & 0.82         & 0.87   &0.67 & 7.72 & 5.81 & \textbf{0.77}\\
\multicolumn{5}{l}{\textbf{GPT 4o}} \\   
\texttt{MT} & 0.83     & 0.56    & 0.61         & 0.87 &0.83 &8.96 & 5.93 & 0.72\\
\texttt{MTOSC}    & 0.84    & 0.69    & 0.79         & 0.95   &0.83 & 8.32 & 5.69 &\textbf{0.82}\\
\multicolumn{5}{l}{\textbf{GPT 5}} \\   
\texttt{MT} & 0.90     & 0.64    & 0.96         & 0.91 &0.5 & 8.81 & 6.66 & 0.85\\
\texttt{MTOSC}      & 0.92     & 0.73    & 1.0         & 0.89 &0.5  & 8.15 & 6.45 & \textbf{0.89}\\
\multicolumn{5}{l}{\textbf{GPToss-20B}} \\   
\texttt{MT} & 0.86     & 0.59    & 0.69         & 0.93 & 0.83 & 7.79 & 5.47 & 0.77\\
\texttt{MT-OSC}  & 0.84   & 0.61    & 0.70         & 0.91   & 0.67 & 7.24 & 5.85 & 0.76\\
\multicolumn{5}{l}{\textbf{GPToss-120B}} \\   
\texttt{MT} & 0.92     & 0.49    & 0.86         & 0.93 & 0.5 & 8.51 & 6.15 & 0.80\\
\texttt{MT-OSC} & 0.94     & 0.66    & 0.82         & 0.89 &0.67 & 7.88 & 6.21 & \textbf{0.83}\\
\hline 
\multicolumn{5}{l}{\textbf{Avg. by dataset}}\\
\texttt{MT} & 0.85 & 0.67 & 0.76 & 0.91 & 0.58 & \textbf{8.67} & 5.84 & 0.80\\
\texttt{MT-OSC} & \textbf{0.86} & \textbf{0.74} & \textbf{0.81} & \textbf{0.92} & \textbf{0.67} & 8.01 & \textbf{5.86} & \textbf{0.83}\\
\bottomrule
\end{tabular}
\caption{Results across datasets and chat models for samples with 6 turns or more using MT-baseline and MT-OSC. (*Spider has a very small sample size of 6 for >=6 turns. Average accuracy per LLM doesn't include Spider due to this reason.)}
\label{tab:llmsdatasetsres}
\end{table*}

\paragraph{Results across Condenser models}
\label{sec:condmodelres}

Table ~\ref{tab:condmodelres} shows results broken down by dataset (sharded BFCL and HumanEval subsets) and chat model (Llama-4-Maverick, GPT 4.1, GPT 4o), for different condenser model choices (Llama-3.3-70B(default), Llama-4-Maverick, GPT4.1, Gemini 2.5 Flash) compared to MT-baseline for each chat model.

\begin{table}[h!]
\centering
\begin{tabular}{p{2.7cm}p{1cm}p{1cm}p{1cm}}
BFCL$^{shrd}$ &&&\\
\hline
\hline
& Llama4-mav & GPT4.1 & GPT4o \\
Condenser &&&\\
\hline
N/A (MT-baseline) & 83.61 & 67.44 & 55.74\\
Llama-3.3-70B& 90.16$\uparrow$ & 80.33$\uparrow$ & 68.85$\uparrow$\\
Llama-4-Mav & 83.61 & 72.13$\uparrow$ & 68.85$\uparrow$\\
GPT 4.1& 86.89$\uparrow$ &  75.41$\uparrow$ & 65.57$\uparrow$\\
Gemini-2.5-flash & 85.25$\uparrow$ & 77.05$\uparrow$ & 63.93$\uparrow$\\
\hline
&&&\\
HumanEval$^{shrd}$ &&&\\
\hline
\hline
& Llama4-mav & GPT4.1 & GPT4o \\
Condenser &&&\\
\hline
N/A (MT-baseline) & 82.72  & 82.72 & 60.49\\
Llama-3.3-70B  & 91.36$\uparrow$ &	90.12$\uparrow$ & 79.01$\uparrow$\\
Llama-4-Mav & 92.59$\uparrow$ & 85.19$\uparrow$ & 77.78$\uparrow$\\
GPT 4.1 & 85.19$\uparrow$ &	85.19$\uparrow$ & 88.89$\uparrow$\\
Gemini-2.5-flash & 82.72 & 80.25 & 72.84$\uparrow$\\
\hline
\end{tabular}
\caption{Across Condenser Models, performance in (\%) of MT-OSC compared to MT-baseline for data samples with at least 6 turns or more in sharded BFCL and HumanEval sets.}
\label{tab:condmodelres}
\end{table}

\subsection{Robustness test results}
\label{sec:robustresults}
Table~\ref{tab:robust} compares the performance between MT-baseline and MT-OSC on top of perturbed dataset samples. Results show that MT-OSC consistency performs on perturbed data as it does on original samples of the same data. 

\begin{table*}[h!]
\centering
\begin{tabular}{l|c|c|c|c|c|c}
\toprule
& \multicolumn{2}{c|}{1. Repetition Infusion} &  \multicolumn{2}{c|}{2. Filler Injection} &  \multicolumn{2}{c}{3. Contextual Diversion} \\
  & MT & MT-OSC & MT & MT-OSC & MT & MT-OSC \\
\hline
GSM$^{shrd}$ & 85.80 & 86.63 &	84.77 & 87.86 &	82.92 &	84.77\\
BFCL$^{shrd}$ & 63.45 & 70.64 & 64.02 & 69.89 & 59.77 & 67.24 \\
HumanEval$^{shrd}$ & 70.81 & 75.89 & 74.40 & 79.17 & 66.67 &	79.63\\
Spider$^{shrd}$ & 67.68 &	70.20 &	63.13 &	66.67 &	57.07 &	61.62\\
refinement$^{mte}$ & 5.67 & 5.66 &	5.7 &	5.64 &	5.01 &	5.00\\
\bottomrule
\end{tabular}
\caption{Comparison of performance between MT and MT-OSC on the three different variants created for robustness testing (1. Repetition Infusion; 2. Filler Injection; 3. Contextual Diversion).}
\label{tab:robust}
\end{table*}

\subsection{Decider analysis}
\label{sec:decider}

\begin{figure}
    \centering
    \includegraphics[width=1\linewidth]{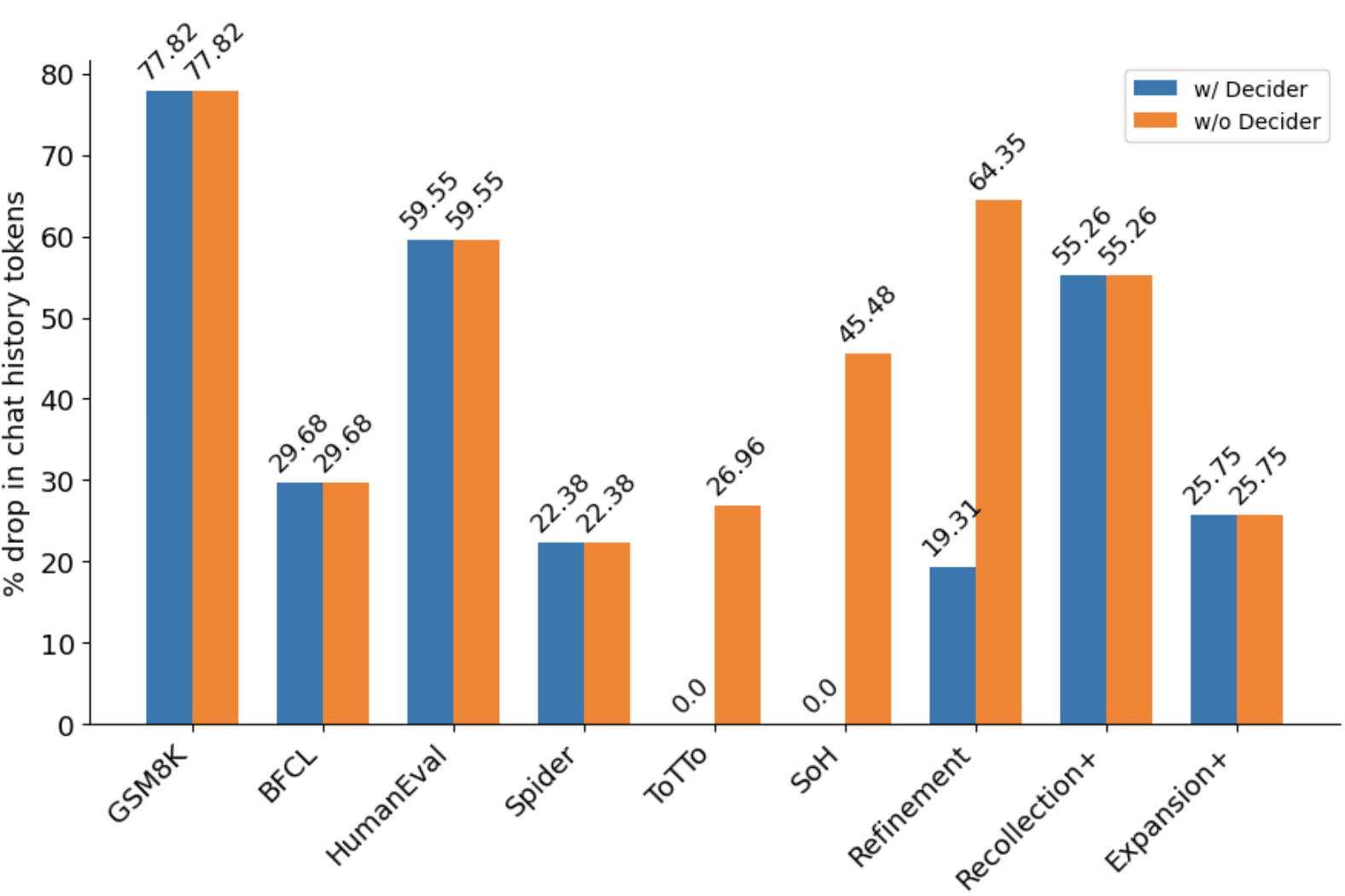}
    \caption{Percentage reduction in chat history tokens in MT-OSC over MT-baseline across datasets with and without Decider component.}
    \label{fig:tokencomp}
\end{figure}

Figure~\ref{fig:tokencomp} shows that chat history tokens are reduced across different datasets when using MT-OSC, both with and without the Decider component, for conversations with six or more turns. 

With the Decider, chat history tokens are reduced by an average of 32.19\%. Without the Decider, the reduction averages 45.58\% across all datasets (even the ones where no sample goes through the condenser in the w/ Decider case). 

For datasets where some samples are condensed in w/ Decider case (BFCL, GSM, HumanEval, Refinement, Recollection+, and Expansion+), with Decider reduction averages 44.56\% and without Decider reduction averages 52.07\%. Note we didn't include Spider as it has only 6 samples with >=6 turns so we excluded it in the calculation due to very low number of samples.

\paragraph{Example}
To complement the Decider ablation results, we share a Sharded ToTTo dataset example (using chat model Llama-3.3-70B) illustrating how default condensation without Decider gate may lose salient multi-turn emphasis signals. In this setting, the final table description depends on information provided across all past turns, including a highlighted-cell view and additional metadata.

\noindent Example user-provided conversation turns:

\noindent\textit{Turn 1:} "I am providing you a table. You must produce a short one-sentence description of the table. The description should be at most 30 words…………..<data>" \\
\textit{Turn 2:} "Here's 10 examples of descriptions of other tables. You can use it as references to understand how to describe the table: <data>"\\
\textit{Turn 3:} "I am now giving you a version of the table with some cells highlights. ……."\\ 
\textit{Turn 4:} "Here's additional metadata about the table. The table\_webpage\_url is ……"\\
\textit{Turn 5:} "*Here's additional metadata about the table. The table\_page\_title is 2012 in AFC."\\
\textit{Turn 6:} "Here's additional metadata about the table. The table\_section\_title is Events list."

In this conversation, earlier turns center more around ``AFC 3", while turns (4–6) provide page and section context that should be incorporated without overriding the highlight focus.

MT-OSC final output without Decider (always condense): ``The table lists events on the 2012 in AFC Wikipedia page under Events list."

MT-OSC final output with Decider (in this case - skip condensation): ``The table highlights AFC 3 event details in the Events list section of the 2012 in AFC Wikipedia page."

This example shows a typical Decider benefit: aggressive condensation may at time, for certain chat models, produce a less specific final description by losing or down-weighting earlier emphasis cues in comparison to cues shared in newer turns; the Decider helps avoid this potential failure mode.

\subsection{Decider parameter sensitivity analysis}
\label{sec:decsens}

\begin{figure*}
    \centering
    \includegraphics[width=1\linewidth]{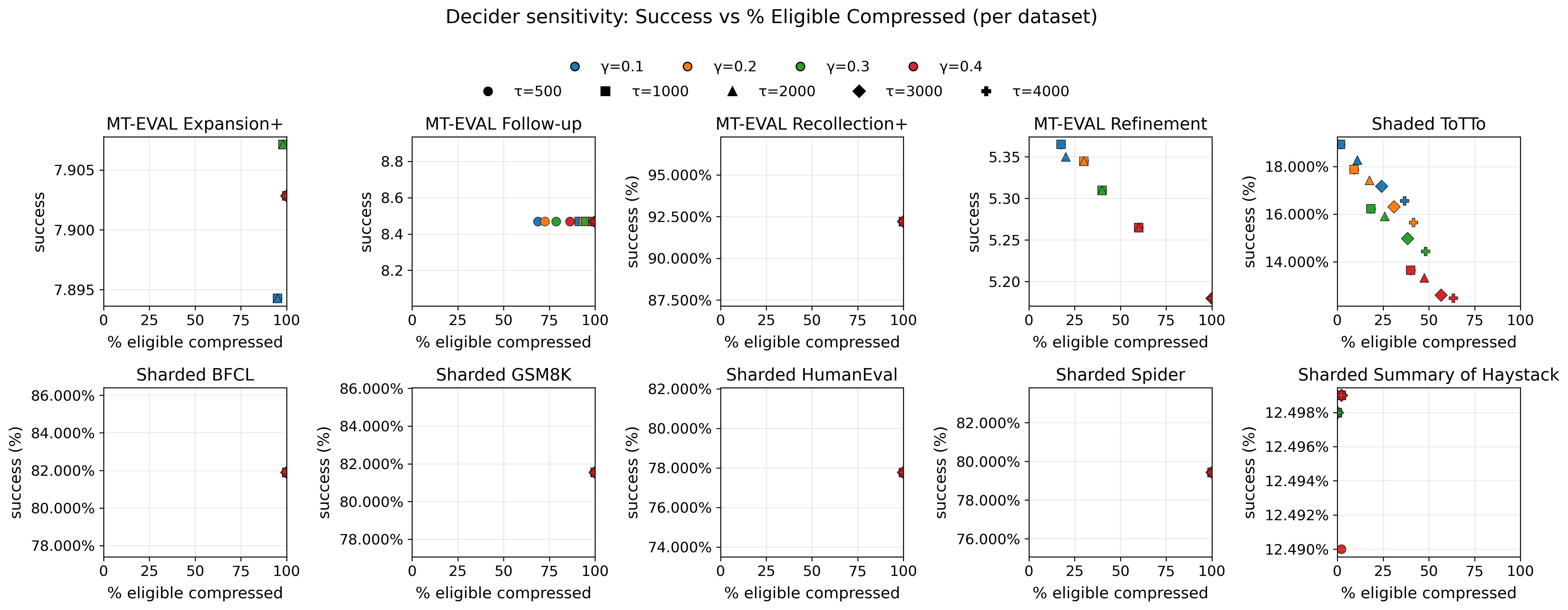}
    \caption{Sensitivity of the decider to $\gamma$ and $\tau$ across datasets. Each panel corresponds to one dataset. Points show, for each ($\gamma$,$\tau$) setting, the end-to-end task success score (y-axis; reported in the dataset’s native units) versus the number of samples that were routed to compression (x-axis). Color indicates $\gamma$ and marker shape indicates $\tau$. Flat panels indicate insensitivity to ($\gamma$,$\tau$), while sloped panels highlight datasets where increased compression changes downstream success.}
    \label{fig:decsens}
\end{figure*}

\paragraph{Sensitivity to decider parameters $(\gamma,\tau)$.}
We sweep $\gamma \in \{0.1, 0.2, 0.3, 0.4\}$ and $\tau \in \{500, 1000, 2000, 3000, 4000\}$ on 10 datasets.
For each setting we report (i) the number of samples routed to compression versus not, and (ii) the resulting end-to-end \emph{success} score (reported in each dataset's native metric).
Figure~\ref{fig:decsens} shows the results.
Overall, \textit{5/10 datasets are insensitive} to $(\gamma,\tau)$ in our sweep (GSM, BFCL, HumanEval, Spider, Recollection+): both the routing decisions and success are unchanged.
Sensitivity is concentrated in \textit{Totto}, where increasing $\tau$ generally increases the number of eligible samples compressed and changes the success score; \textit{Refinement} has a similar trend as ToTTo but the change in success score is much smaller. \textit{Expansion+} show very small differences; \textit{Summary of Haystack} is almost constant with only a minor change at $\gamma=0.4$; \textit{Follow-up} is constant since it has <=6 turns, ultimately not getting compressed due to turn limit (MT-OSC at w=4 applies to a chat >=6 turns).

These results indicate that the decider hyperparameters primarily control compression aggressiveness on a subset of datasets, while some datasets remain robust, a design allowing for compressing but avoiding risk of information loss which some of these datasets samples portray.  

\section{Condenser and Summarizer examples}
\label{sec:examples}

\begin{figure*}
    \centering
    \includegraphics[width=1\linewidth]{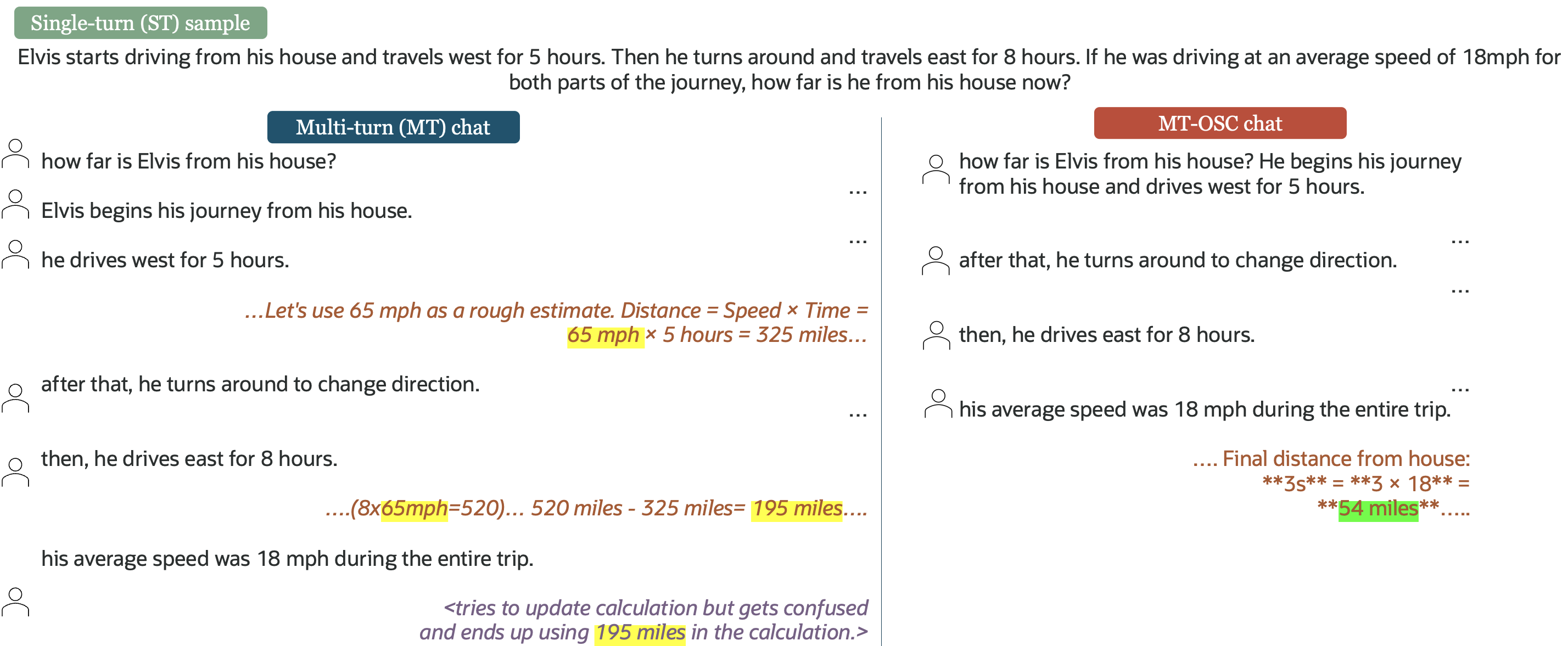}
    \caption{Example showing MT-baseline vs MT-OSC, where MT-OSC handles incorrect assumptions made by the model earlier in a chat.}
    \label{fig:example}
\end{figure*}

Figure~\ref{fig:example} presents a sample from the GSM8K$^{shrd}$ dataset in which the MT-baseline run results in incorrect assumptions and an incorrect final answer. In contrast, MT-OSC successfully addresses these issues and produces the correct solution.

The below example show a sample chat where assistant makes assumptions not provided by the user. The simple summary fails to separate user shared context from assistant shared context. The MT-OSC Condenser module makes a better separation. Keep the context separate is useful so future turns of the conversation are not confused by potential incorrect assumptions made by the assistant model earlier in the chat.

\paragraph{Example original chat:}
This example shows first four turns on a chat sample from GSM8K$^{shrd}$.

- \textbf{USER}: \textit{Q: how far is Elvis from his house?\textbackslash{n}A:}

- \textbf{ASSISTANT}: \textit{To answer this question, I need more information. The question doesn't provide enough context or details about Elvis's location or the location of his house.\textbackslash{n}\textbackslash{n}Could you please provide more information, such as:\textbackslash{n}\textbackslash{n}* Where is Elvis currently located?\textbackslash{n}* Where is his house located?\textbackslash{n}* Is there a specific distance or direction mentioned?\textbackslash{n}\textbackslash{n}With more context, I can try to help you solve the problem and estimate the distance between Elvis and his house.}

- \textbf{USER}: \textit{Q: Elvis begins his journey from his house.\textbackslash{n}A:}

- \textbf{ASSISTANT}: \textit{With this new information, we can start fresh.\textbackslash{n}\textbackslash{n}Since Elvis begins his journey from his house, that means he is currently at his house.\textbackslash{n}\textbackslash{n}So, to answer the original question: "How far is Elvis from his house?"\textbackslash{n}\textbackslash{n}The answer would be: Elvis is 0 miles (or 0 kilometers) away from his house, since he is starting from there.}

- \textbf{USER}: \textit{Q: he drives west for 5 hours.\textbackslash{n}A:}

- \textbf{ASSISTANT}: \textit{New development!\textbackslash{n}\textbackslash{n}Since Elvis drives west for 5 hours, we need to consider the distance he covers during this time. However, we don't know his exact speed.\textbackslash{n}\textbackslash{n}Let's assume a moderate speed of 60 miles per hour (mph) for simplicity. This is just an estimate, as the actual speed might vary.\textbackslash{n}\textbackslash{n}Distance = Speed × Time\textbackslash{n}= 60 mph × 5 hours\textbackslash{n}= 300 miles\textbackslash{n}\textbackslash{n}So, Elvis drives 300 miles west from his house.\textbackslash{n}\textbackslash{n}To answer the original question: "How far is Elvis from his house?"\textbackslash{n}\textbackslash{n}The answer would be: Elvis is approximately 300 miles away from his house, in a westerly direction.}

- \textbf{USER}: \textit{'Q: after that, he turns around to change direction.\textbackslash{n}A:}

- \textbf{ASSISTANT}: \textit{Plot twist!\textbackslash{n}\textbackslash{n}Since Elvis turns around to change direction, that means he is now heading in the opposite direction, which is east.\textbackslash{n}\textbackslash{n}He was previously 300 miles west of his house. Now, he starts driving east, but we don't know for how long or how far.\textbackslash{n}\textbackslash{n}The distance between Elvis and his house remains the same for now: 300 miles. However, as he starts driving east, this distance will begin to decrease.\textbackslash{n}\textbackslash{n}We'll need more information about his speed and the time he drives east to determine the new distance between Elvis and his house.}

\paragraph{Simpler summarizer output:}
This example shows the output of the chat shared above when passed through the simple summarizer.

- \textbf{USER}: \textit{The conversation starts with a question about the distance between Elvis and his house. It is then clarified that Elvis begins his journey from his house, so he is 0 miles away. He then drives west for 5 hours, covering an estimated distance of 300 miles at a moderate speed of 60 mph. After that, he turns around to change direction, heading east, but the time and distance driven east are not specified.}

- \textbf{ASSISTANT}: \textit{The assistant responds by initially requesting more context about Elvis's location. After each update, the assistant recalculates the distance between Elvis and his house: first it's 0 miles when he starts at home, then approximately 300 miles after driving west for 5 hours at 60 mph, and finally this distance begins to decrease as he heads back east but the exact new distance is unknown without more information on his speed and time driven east.}

\paragraph{MT-OSC Condenser output:}
This example shows the output of the Condenser component for the original chat shared above.

- \textbf{USER}: \textit{how far is Elvis from his house? He begins his journey from his house, drives west for 5 hours and then turns around to change direction.}

- \textbf{ASSISTANT}: \textit{Elvis is initially 0 miles away from his house. After driving 300 miles west, he is approximately 300 miles away. When he turns around, the distance remains 300 miles until he starts driving east.}

\end{document}